\documentclass[10pt,twocolumn,letterpaper]{article}

\usepackage{cvpr}              % To produce the CAMERA-READY version
\definecolor{cvprblue}{rgb}{0.21,0.49,0.74}
\usepackage[pagebackref,breaklinks,colorlinks,allcolors=cvprblue]{hyperref}

\usepackage{graphicx}
\usepackage{multirow}
\usepackage{subcaption}
\usepackage[table]{xcolor}
\usepackage[utf8]{inputenc} % allow utf-8 input
\usepackage[T1]{fontenc}    % use 8-bit T1 fonts
\usepackage{hyperref}       % hyperlinks
\usepackage{url}            % simple URL typesetting
\usepackage{booktabs}       % professional-quality tables
\usepackage{amsfonts}       % blackboard math symbols
\usepackage{nicefrac}       % compact symbols for 1/2, etc.
\usepackage{microtype}      % microtypography
\usepackage{booktabs} % 好看的三线表
\usepackage{tabularx}
\usepackage{stfloats}
\usepackage{colortbl}
\usepackage{svg}\usepackage{svg}
\usepackage{flushend}

\usepackage{fontawesome5}
\usepackage{colortbl}
\usepackage{tikz}
\usepackage{ulem}
\usepackage[most]{tcolorbox}
\usepackage{fancyvrb}
\usepackage{fvextra}

\usepackage{wrapfig}

\definecolor{cityblue}{RGB}{128, 159, 225}
\definecolor{citypink}{RGB}{227, 108, 194}
\definecolor{w_blue}{RGB}{237, 241, 253}
\usepackage{array,tabularx}
\newcolumntype{C}{>{\centering\arraybackslash}X}
\definecolor{colorbest}{RGB}{252,187,161}
\definecolor{colorsecond}{RGB}{254,224,210}
\definecolor{colorthird}{RGB}{255,245,240}
\definecolor{dataconstruction}{RGB}{109,153,255}

\usepackage{caption}
\title{
RealRestorer: Towards Generalizable Real-World Image Restoration with Large-Scale Image Editing Models}

\author{
    \normalsize
    Yufeng Yang$^{1,2}$ \quad Xianfang Zeng$^{2,\dag}$ \quad Zhangqi Jiang$^{2}$ \quad Fukun Yin$^{2}$ \quad Jianzhuang Liu$^{3}$ \quad Wei Cheng$^{2}$ \\
    \normalsize
    Jinghong Lan$^{2}$ \quad Shiyu Liu$^{2}$ \quad Yuqi Peng$^{3}$ \quad Gang Yu$^{2,\ddag}$ \quad Shifeng Chen$^{3,4,\ddag}$ 
    \\[0.4em]
    \normalsize
    $^{1}$Southern University of Science and Technology \quad $^{2}$StepFun \\
    \normalsize
    $^{3}$Shenzhen Institutes of Advanced Technology, Chinese Academy of Sciences \quad $^{4}$Shenzhen University of Advanced Technology
    \\[0.4em]
    \textcolor{cityblue}{\normalsize 
    \raisebox{-0.2\height}{\includegraphics[height=0.45cm]{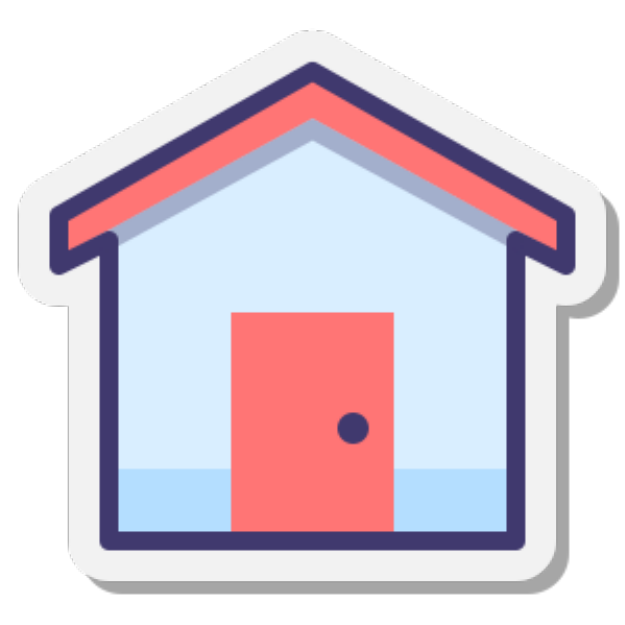}}~\href{https://yfyang007.github.io/RealRestorer/}{\textbf{Project Page}}
    \quad
    \raisebox{-0.2\height}{\includegraphics[height=0.45cm]{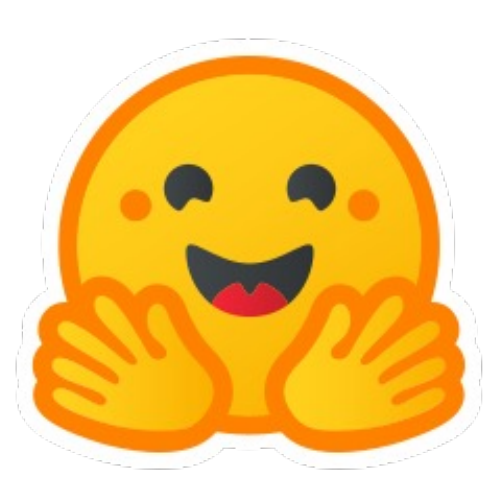}}~\href{https://huggingface.co/RealRestorer/RealRestorer}{\textbf{Models}}
    \quad
    \raisebox{-0.2\height}{\includegraphics[height=0.45cm]{img/huggingface_logo.pdf}}~\href{https://huggingface.co/datasets/RealRestorer/RealIR-Bench}{\textbf{RealIR-Bench}}
    \quad
    \raisebox{-0.2\height}{\includegraphics[height=0.45cm]{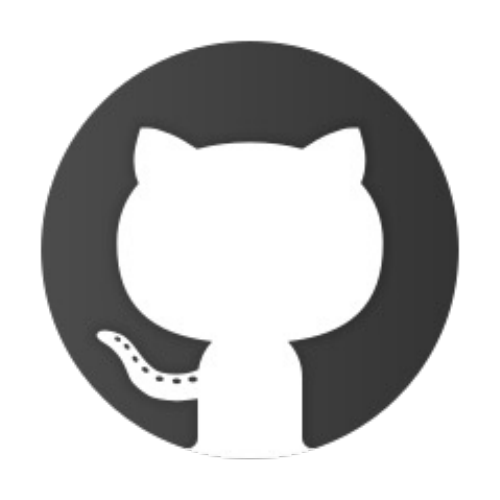}}~\href{https://github.com/yfyang007/RealRestorer}{\textbf{Code}}
    }
}

\begin{document}

\twocolumn[{
  \renewcommand\twocolumn[1][]{#1}
  \maketitle
  \begin{center}
  \vspace{-5ex}
  \includegraphics[width=\textwidth]{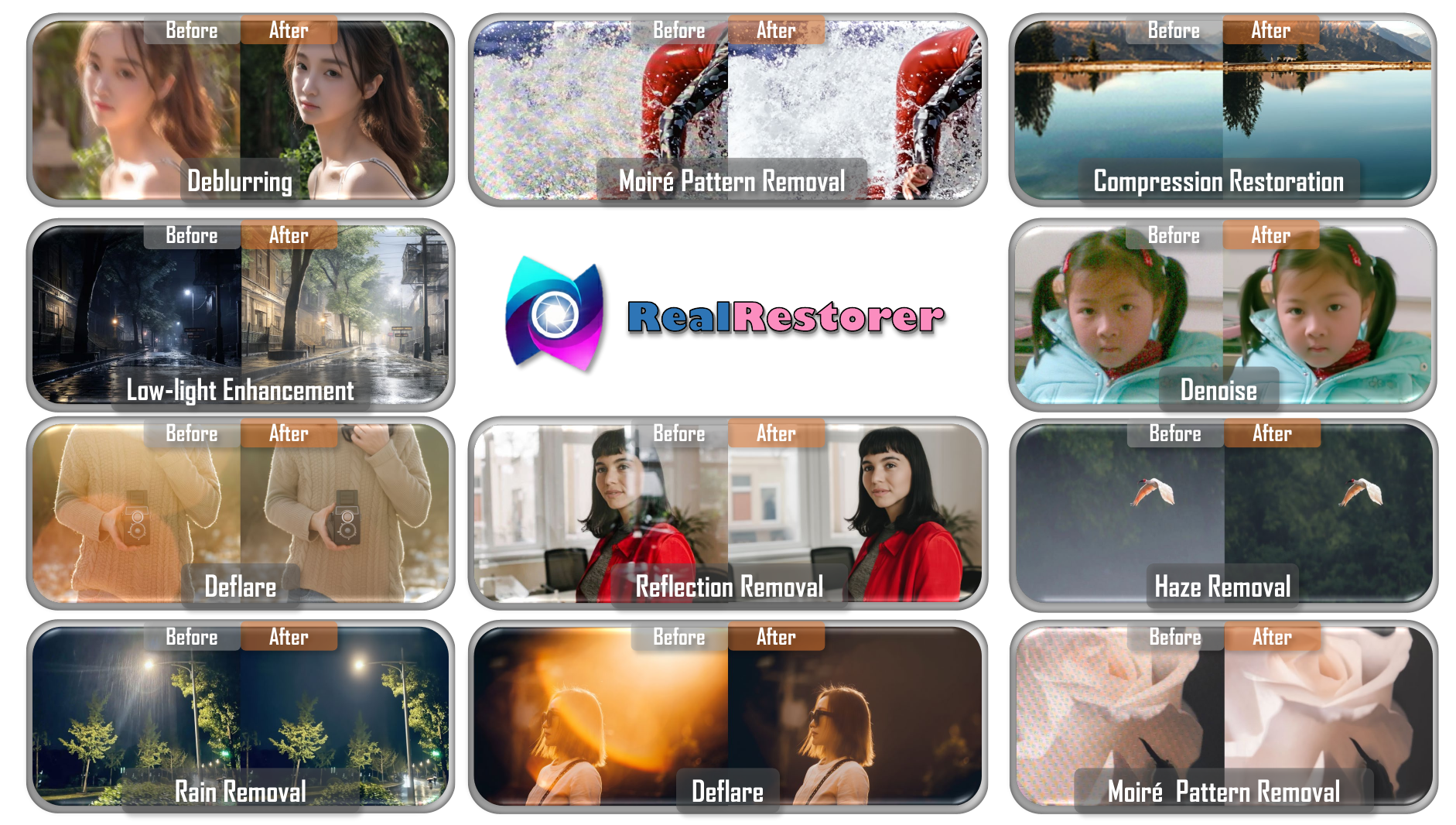}
  \captionof{figure}{\small RealRestorer effectively restores diverse real-world image degradations, including deblurring, moiré pattern removal, compression restoration, reflection removal, hazing removal, rain removal, deflare, and low-light enhancement.}
  \label{fig:teaser}
  \end{center}
}]

{\let\thefootnote\relax\footnotetext{\noindent$\dag$ leads this project; \ddag Corresponding authors.}}

\begin{abstract}
% Image restoration under real-world degradations is critical for downstream tasks such as autonomous driving and object detection. However, existing restoration models are often limited by the scale and distribution of their training data, resulting in poor generalization to real-world scenarios. Recently, large-scale image editing models have shown strong generalization ability in restoration tasks, especially for closed-source models like Nano Banana Pro, which can restore images while preserving consistency. Nevertheless, achieving such performance with those large universal models requires substantial data and computational costs.
% To address this issue, we construct a large-scale dataset covering nine common real-world degradation types, train a strong open-source image editing model, and introduce RealIR-Bench, which contains 464 real-world degraded images and tailored evaluation metrics focusing on degradation removal and consistency preservation. Extensive experiments show that our model ranks first among open-source methods and third overall, achieving near state-of-the-art performance.

Image restoration under real-world degradations is critical for downstream tasks such as autonomous driving and object detection. However, existing restoration models are often limited by the scale and distribution of their training data, resulting in poor generalization to real-world scenarios. Recently, large-scale image editing models have shown strong generalization ability in restoration tasks, especially for closed-source models like Nano Banana Pro, which can restore images while preserving consistency. Nevertheless, achieving such performance with those large universal models requires substantial data and computational costs. To address this issue, we construct a large-scale dataset covering nine common real-world degradation types and train a state-of-the-art open-source model to narrow the gap with closed-source alternatives. Furthermore, we introduce RealIR-Bench, which contains 464 real-world degraded images and tailored evaluation metrics focusing on degradation removal and consistency preservation. Extensive experiments demonstrate our model ranks first among open-source methods, achieving state-of-the-art performance.

\end{abstract}

% \section{Introduction}\label{sec:intro}

\section{Introduction}
\label{sec:intro}
Image restoration~\cite{liang2021swinir,gunturk2018image,lehtinen2018noise2noise,zamir2021multi,li2023lsdir} aims to recover high-quality images from degraded observations and serves as a fundamental building block for downstream applications such as autonomous driving~\cite{hu2023planning,caesar2020nuscenes}, remote sensing~\cite{wang2024improved}, detection~\cite{hu2018relation,joseph2021towards}, and 3D reconstruction~\cite{yao2018mvsnet}. However, real-world images often suffer from diverse and co-existing degradations~\cite{liang2022efficient,huang2020real,flusser2015recognition,damera2000image,zhang2021designing,dufaux2000automatic,jiang2021rain,liu2021benchmarking,guo2023low,he2021moire,hermann2012periodic,al2015investigating,tong1996photo,benz2017flare}, including blur, rain, noise, low-light, moiré patterns, haze, compression artifacts, reflection, and flare. This complexity goes beyond the single degradation and single model paradigm.

To address this, recent all-in-one restoration methods~\cite{li2022all,zamfir2025complexity,potlapalli2023promptir,huang2024wavedm} attempt to handle multiple degradations within a unified framework. Nevertheless, they often rely on a limited set of synthetic degradation distributions, while collecting large-scale real degraded-clean pairs remains expensive and difficult. As a result, these models can generalize poorly to real-world scenarios. In parallel, large image editing models trained on massive editing datasets have recently demonstrated strong restoration capabilities~\cite{zuo2025nano}, such as Nano Banana Pro~\cite{team2023gemini} and GPT-Image-1.5~\cite{gpt4o20250325}. However, these models are typically trained with closed-source data and compute, which makes them hard to reproduce and limits their utility for the research community. Despite this, leveraging the strong priors learned by image editing models provides a promising path to overcome the key limitation of traditional restoration approaches.

However, conventional restoration datasets often focus on a narrow degradation distribution that is not representative of real-world conditions. Evaluation protocols that emphasize only reference-based metrics further exacerbate this issue, as they may not reflect perceptual quality, robustness across diverse degradations, or detail consistency in real scenes.

To bridge these gaps, we design a comprehensive degradation synthesis pipeline to generate high-quality training data, aiming to narrow the gap between synthetic and real-world degradations. Based on this dataset, we fine-tune an open-source image editing model \textbf{RealRestorer} across nine restoration tasks, and further introduce a new benchmark \textbf{RealIR-Bench} to evaluate restoration performance under real-world degradations.

In summary, our contributions are threefold:
\begin{itemize}
\item We develop \textbf{RealRestorer}, an open-source real-world image restoration model that sets a new state of the art and achieves performance highly comparable to closed-source systems. We will release the model to facilitate future research in real-world restoration.

\item We propose a data generation pipeline to produce high-quality restoration training data with diverse and representative degradations. This pipeline provides a valuable resource for developing more robust restoration models.

\item We develop a new benchmark, \textbf{RealIR-Bench}, grounded in real-world cases, to evaluate both degradation restoration and consistency preservation. By addressing the lack of reliable evaluation protocols for real-world restoration, it enables more authentic and comprehensive assessment of restoration models.
\end{itemize}
% \section{Related Work}\label{sec:related}
\section{Related Work}\label{sec:related}
\subsection{Single-Degradation Restoration}
% Single-degradation restoration methods typically focus on removing one specific type of degradation under constrained scenarios. With the development of deep learning, numerous works~\cite{Nah_2017_CVPR,li2018benchmarking,cai2023retinexformer,zhao2025reversible, huang2024wavedm} have achieved impressive performance on tasks such as deblurring, haze removal, compression restoration, low-light enhancement, deflare, and reflection removal. However, due to the task-specific modeling assumption, models trained for a single degradation often fail when multiple degradations are present in real-world images, which significantly limits their practicality and robustness in real applications.
Single-degradation restoration methods typically focus on removing one specific type of degradation under constrained and well-defined scenarios. With the rapid development of deep learning, numerous works~\cite{Nah_2017_CVPR,li2018benchmarking,cai2023retinexformer,zhao2025reversible,huang2024wavedm} have achieved impressive performance on individual tasks such as deblurring, haze removal, low-light enhancement, deflare, and reflection removal. These approaches often rely on carefully designed architectures and degradation-specific priors, enabling strong performance.

However, most single-degradation models are built upon task-specific assumptions, where the degradation type is predefined and relatively homogeneous, which makes models trained for a single degradation tend to generalize poorly and may even introduce secondary artifacts when encountering unseen or compound degradations.

Moreover, many existing methods are trained and evaluated primarily on synthetic datasets with simplified degradation models, which may not faithfully represent the complexity of real-world data distributions. This gap between synthetic training data and real-world testing scenarios further limits their robustness and practical applicability. Consequently, while single-degradation methods achieve strong performance on benchmark datasets, their effectiveness in real-world applications remains constrained.

\subsection{All-in-One Image Restoration}
All-in-one approaches~\cite{li2025foundir,lin2024diffbir,potlapalli2023promptir,luo2023controlling,li2022all,cui2025adair,zamfir2025complexity, guo2024onerestore} aim to handle multiple degradations within a unified network by balancing shared representations and task-specific components. Nevertheless, many of these methods still rely heavily on synthetic datasets with limited and overly simplified degradation patterns. Such a narrow training distribution often results in weak robustness and poor generalization to real-world degradations, where corruption characteristics are diverse, complex, and domain-dependent.

Meanwhile, large diffusion or flow-matching image editing models~\cite{lipman2022flow,esser2024scaling,peebles2023scalable,rombach2022high} have recently demonstrated strong semantic priors for image enhancement and restoration. Trained on massive image–text pairs, these image editing models~\cite{labs2025flux,liu2025step1x,wu2025qwenimagetechnicalreport, LongCat-Image} can leverage semantic conditioning and often generalize better to real-world data than small specialized restoration networks. Therefore, transferring and exploiting the priors of large image editing models provides a promising direction for building restoration systems with stronger real-world generalization.

Motivated by this observation, we develop a high-quality and realistic degradation synthesis pipeline covering nine major degradations and use it to fine-tune open-source image editing models for robust real-world restoration while maintaining strong content consistency. Furthermore, to evaluate real-world restoration performance in the absence of clean references, we curate a benchmark of 464 real images spanning nine single-degradation categories, and propose new evaluation metrics that measure both degradation removal ability and consistency with the input content. Based on the proposed dataset and metrics, our fine-tuned model achieves state-of-the-art performance among open-source methods and is competitive with closed-source systems, while qualitative results further demonstrate strong generalization to real-world degradations.

% \section{\ourdataset : Paired Multi-Person Dataset Construction}\label{sec:dataset}

\section{RealRestorer}\label{sec:method}
\subsection{Data Construction}

\begin{figure*}[t]
    \centering
    \setlength{\abovecaptionskip}{2pt}
    \includegraphics[width=0.98\linewidth]{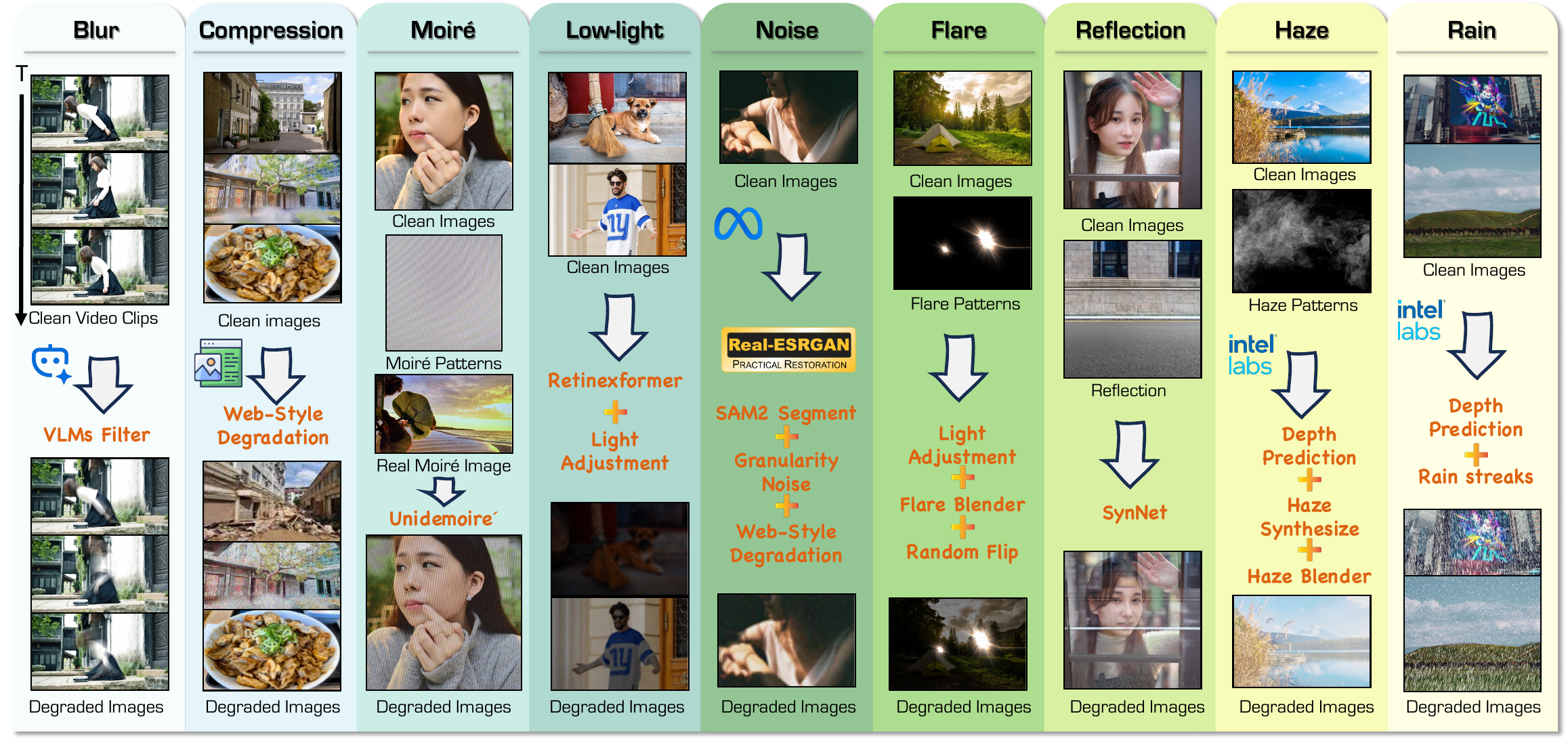}
    \vspace{0.5ex}
    \caption{\small Overview of our large-scale Synthetic Degradation Data pipeline. 
    We construct nine representative degradation types, including blur, compression artifacts, moiré patterns, low-light, noise, flare, reflection, haze, and rain. 
    Compared with previous synthetic-only pipelines, our upgraded framework incorporates granular noise modeling, segment-aware perturbations, and web-style degradation processes, significantly narrowing the gap between synthetic and real-world distributions. 
    This comprehensive pipeline enables more robust and generalizable restoration learning.}
    \vspace{-3ex}
    \label{fig:pipeline}
\end{figure*}

Existing image restoration datasets~\cite{li2025foundir,guo2024onerestore} often rely on a single degradation model to synthesize degraded images and use a fixed composition strategy to explicitly disentangle degradation features for representation learning. These modeling approaches are effective for specific degradation settings. However, in real-world scenarios, degradations are far more complex and diverse. Simple synthetic degradation models are usually insufficient to approximate real degradation distributions, and they are often not robust enough for large-scale training that aims at strong generalization.

To address this limitation, we develop a new dataset collection pipeline that produces more realistic degradation patterns while keeping the paired clean images highly consistent with their degraded counterparts. 

In general, we adopt two main ways to obtain high-quality paired data for image restoration of nine tasks:

\noindent\textbf{Synthetic Degradation Data}: Start from clean images and synthesize degradations. This approach is highly scalable as long as sufficient clean images can be collected from the internet. However, even with increasingly sophisticated degradation synthesis, it remains challenging to fully capture the diversity and complexity of real-world degradations. Nevertheless, such synthetic data can still be valuable, as it provides a convenient way to transfer general image editing priors to image restoration models and helps them acquire foundational restoration knowledge. We leverage several powerful open-source models to support the synthetic data generation process, including SAM-2~\cite{ravi2024sam}, and MiDaS~\cite{Ranftl2022}. These models are used to filter unsuitable samples and provide essential structural and geometric information required for realistic degradation synthesis, such as semantic masks and depth cues.

In our pipeline, to ensure high data quality, we employ the Vision-Language Models (VLMs) and quality assessment models~\cite{minderer2023scaling} to filter out low-quality or unsuitable images like watermarked images. After forming pairs, we further examine the degree of degradation alignment between the degraded and restored images to ensure that the degradation patterns are learnable from the paired data. Specifically, the synthetic pairing data construction is illustrated as follows.

\textbf{Blur:}
The motion blur dataset is primarily synthesized using temporal averaging over video clips to simulate realistic motion trajectories. Both the target and source images are filtered to ensure consistent blur patterns. In addition, web-style degradation, including common blur operations, such as Gaussian blur and standard motion blur, is incorporated to better approximate real-world motion blur characteristics.

\textbf{Compression Artifacts:}
We simulate compression artifacts using JPEG compression and image resizing to approximate common web compression effects. In addition to standard JPEG degradation, we also incorporate web-style compression processes to better reflect the wide range of compression artifacts found in online images.

\textbf{Moiré Patterns:}
Following UniDemoiré~\cite{yang2025unidemoire}, we generate 3,000 moiré patterns at multiple scales and randomly fuse one to three patterns into clean images. This strategy substantially improves the diversity and generalization capability of the model for moiré pattern removal.

\textbf{Low-Light:}
We simulate low-light conditions by applying brightness attenuation and gamma correction to reduce pixel intensity. Moreover, we train a separate model~\cite{cai2023retinexformer} using paired datasets such as LOL~\cite{yang2021sparse} and LSRW~\cite{hai2023r2rnet}, reversing the low-exposure and high-exposure image pairs. This trained model is then applied to clean images to better mimic realistic low-light distributions.

\textbf{Noise:}
We adopt web-style degradation as the primary noise synthesis pipeline. Compared with the degradation strategy used in Real-ESRGAN~\cite{wang2021realesrgan}, we further introduce granular noise for web images. Additionally, we incorporate segment-aware noise, which significantly improves performance on real-world denoising tasks.

\textbf{Flare:}
We collect more than 3,000 glare patterns and adapt them to clean images for realistic blending. In addition, random horizontal and vertical flipping is applied to further enhance the diversity of the generated data pairs.

\textbf{Reflection:}
For reflection degradation synthesis, we collect two sources of clean images. The first source mainly consists of portrait images, which are treated as transmission layers. The second source contains diverse scenes with human faces, which are used as reflection layers. To increase the diversity of the paired data, we randomly swap a few portions of the image pairs, using human portraits as reflection layers instead of transmission layers.
The overall synthesis pipeline follows SynNet~\cite{Wen_2019_CVPR}.

\textbf{Haze:}
We synthesize hazy images based on the classic atmospheric scattering model by estimating depth from clean images and generating fog accordingly~\cite{he2010single}. To better simulate real haze, we collect nearly 200 haze patterns and randomly blend them with the synthesized haze, making the results closer to real-world haze distributions.

\textbf{Rain:}
To synthesize realistic rain degradation, we not only add rain streaks but also incorporate splashes and simulate physical effects such as perspective distortion and droplet sputtering. Furthermore, we collect 200 real rain patterns and randomly blend them into clean images to enhance diversity and realism. Besides, we also adopt the rain category from the FoundIR dataset~\cite{li2025foundir}, which contains about 70K paired samples.

\noindent\textbf{Real-World Degradation Data}: 
Collect real degraded images and generate corresponding clean images by removing degradations using high performance restoration models. Compared with synthetic pairing, this approach is more likely to preserve the true degradation statistics of real-world data, enabling restoration models trained on such pairs to generalize better to real scenarios.
To bridge the gap between synthetic and real-world degradations, we collect real degraded images from the web and pair them with high-quality references. 

During web data collection, we first employ the CLIP model~\cite{radford2021learning} to filter images based on degradation-related semantic cues. While this approach effectively removes a portion of irrelevant samples, it still introduces noisy cases, such as watermarked images or visually similar but non-degraded content. To further refine the dataset, we apply a watermark detection filter and leverage Qwen3-VL-8B-Instruct~\cite{qwen3technicalreport} to assess and verify the degree of degradation. After generating clean references using high-performance image generation models, we further examine the consistency of the paired data by employing low-level metrics to detect potential content shifts. A subset of the filtered pairs is then manually reviewed to ensure that the degradation type and severity are properly aligned between degraded inputs and their corresponding clean references. These curated real-world degradation samples enable the model to better adapt its parameters to realistic data distributions. Such adaptation helps the model converge more effectively toward real-world scenarios, consistent with prior findings in large-scale generative modeling~\cite{podell2023sdxl,team2025zimage,LongCat-Image}.

Additional details and qualitative demonstrations are provided in Appendix~\ref{sec:Data}.

\begin{figure*}[thbp]
    \centering
    \setlength{\abovecaptionskip}{2pt}
    \includegraphics[width=0.98\linewidth]{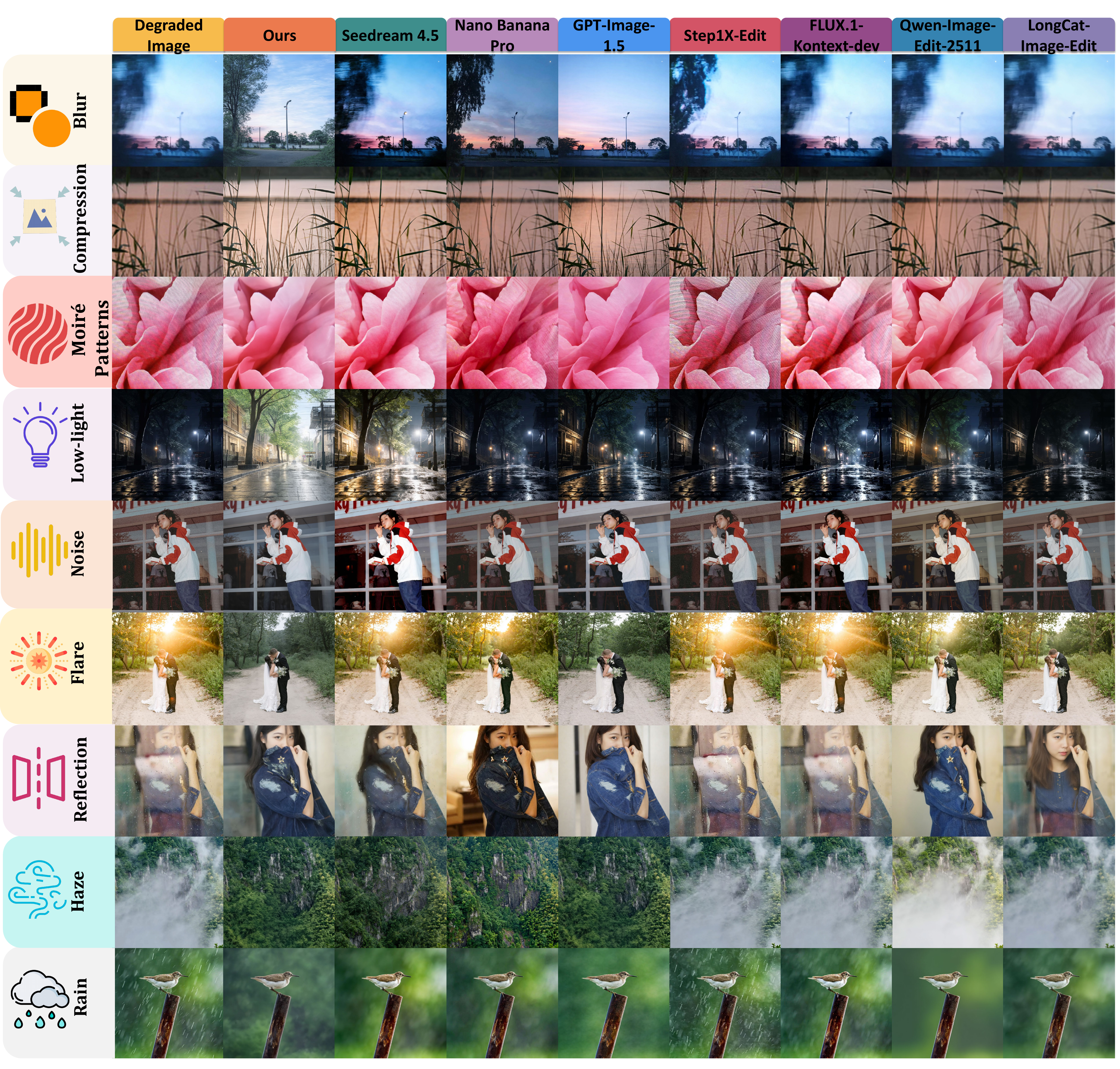}
    \vspace{0.5ex}
    \caption{\small Comparison with state-of-the-art image editing models across nine real-world degradations, including blur, compression artifacts, moiré patterns, low-light, noise, flare, reflection, haze, and rain. We compare our method with large-scale image editing models, such as Seedream 4.5, Nano Banana Pro, GPT-Image-1.5, Step1X-Edit, FLUX.1-Kontext-dev, Qwen-Image-Edit-2511, and LongCat-Image-Edit.}
    \vspace{-3ex}
    \label{fig:comp}
\end{figure*}

\subsection{Method and Training Strategy}
We fine-tune the base model Step1X-Edit~\cite{liu2025step1x} built on a large Diffusion in Transformer (DiT) backbone~\cite{peebles2023scalable}, which is effective for generation. It is equipped with QwenVL~\cite{bai2025qwen3} as a text encoder that injects high-level semantic extraction into the DiT denoising pathway. Inside the diffusion network, a dual-stream design is used to jointly process semantic information together with noise and the conditional input image. The reference image and output image are both encoded into latent space by Flux-VAE~\cite{flux2024}. During training, all the components are initialized from the officially released checkpoint of Step1X-Edit, and we freeze the Flux-VAE
and text encoder, only fine-tune the DiT.
Starting from the original image editing model, we fine-tune on nine restoration tasks in two stages: a Transfer-training stage for large-scale restoration transfer and a Supervised
Fine-tuning stage for constraining the manifold of the final model distribution. 

\noindent\textbf{Transfer Training Stage:} In the first stage, we use synthetic paired data to transfer high-level knowledge and priors from image editing to image restoration. Since we initialize from
a pretrained backbone, we eschew progressive resolution
schedules~\cite{liu2025step1x} for training. Instead, we adopt a high-resolution setting of 
1024×1024 throughout the entire training process. The learning rate is kept constant at $1e^{-5}$, and the global batch size is set to 16. Since most of our training data has a resolution higher than 
1024×1024, no additional upsampling is required, which helps preserve fine-grained details and maintain training stability. For each degradation of nine, we adopt single and fixed prompts, which are also the same for the second training stage. For multi-task learning, we adopt an average sampling ratio across all tasks during training. After several steps of transfer training, \textbf{RealRestorer} begins to exhibit signs of knowledge transfer from high-level image editing tasks to image restoration tasks, which is insufficient in the base model.

Although \textbf{RealRestorer} gradually acquires the basic capability to handle simple degradation patterns across all nine tasks, its ability to distinguish and model diverse real-world degradation patterns remains limited. In particular, the model still struggles to capture fine-grained details in complex scenarios. In some cases, noticeable artifacts are present, and the model fails to respond effectively to certain types of degradations. This observation motivates us to introduce a second training stage aimed at improving generalization and restoration quality under real-world degradation scenarios. Moreover, we observe that different task types exhibit distinct learning dynamics and require varying training durations. Therefore, we select a balanced trade-off checkpoint at the end of the first stage to preserve both generation capability and cross-task generalization.

\noindent\textbf{Supervised Fine-tuning Stage:} For the second training stage, we incorporate real-world degradation data to further enhance restoration fidelity and improve generalization under real-world degradation scenarios~\cite{wu2025qwenimagetechnicalreport,team2025zimage, LongCat-Image}. Compared with the first stage, this stage emphasizes adaptation to complex and authentic degradation patterns.
We adopt a cosine annealing learning rate schedule, where the learning rate is gradually decayed to zero, using the same initial learning rate as in the first stage. This smooth decay strategy stabilizes the transition between training stages and encourages the model to progressively adapt to the real-to-clean paired data. By gradually reducing the optimization step size, the model is guided to converge toward a parameter configuration that better aligns with the distribution represented by the high-quality real-world dataset, thereby improving restoration fidelity and robustness under realistic degradations.

Importantly, instead of completely replacing synthetic data, we adopt a \textbf{Progressively-Mixed} training strategy, which retains a small proportion of synthetic paired samples during the second stage. \textbf{RealRestorer} is first exposed to diverse synthetic degradations to build broad generalization, and then gradually adapted to real-world degradations while maintaining exposure to synthetic distributions. Such a hybrid curriculum helps prevent overfitting to specific real degradation patterns and preserves cross-task robustness. More detailed discussions and quantitative analyses of this training strategy are provided in the ablation study.
In addition, we introduce a web-style degradation data augmentation strategy throughout the training process to enhance robustness to images collected from the web. Such images typically suffer from low visual quality, compression artifacts, and other degradations. By simulating these practical degradation patterns during training, the model becomes better equipped to handle real-world inputs and produce better restoration results under challenging conditions.

Throughout the two-stage training process, we select the intermediate checkpoint with the best generalization capability to maintain a balanced performance across multiple tasks and ensure strong overall performance of the final model.
All our experiments are conducted on 8 NVIDIA H800
GPUs. More implementation details can be found in Appendix~\ref{sec:train}.

% \section{\ourbenchmark : Comprehensive ID Customization Evaluation}\label{sec:benchmark}

\section{Benchmark and Evaluation}

\subsection{RealIR-Bench}

Traditional image restoration benchmarks primarily focus on single-degradation tasks with synthetic corruptions or limited degradation patterns, which makes them insufficient for evaluating model performance in real-world applications~\cite{guan2025weatherbench,guo2024onerestore,rajagopalan2024gendegdiffusionbaseddegradationsynthesis,li2025foundir}. Such benchmarks often fail to capture the complexity, diversity, and unpredictability of degradations encountered in practical scenarios.

To properly evaluate restoration performance under real-world degradations, we construct a new benchmark composed entirely of internet-sourced, naturally degraded images. The proposed benchmark spans nine common restoration tasks and covers a wide range of degradation types frequently observed in real-world photography, including blur, rain, noise, low-light, moiré patterns, haze, compression artifacts, reflection, and flare, which collectively represent the most common forms of real-world image degradation.
To preserve the authentic real-world degradation distribution, we directly curate images from web sources rather than synthesizing degradations. We further conduct manual filtering to ensure both quality control and diversity across degradation types, scene content, and severity levels. This human-in-the-loop curation process helps preserve realistic degradation characteristics while avoiding overly biased, repetitive, or low-quality samples. By combining automatic collection with manual verification, we ensure that the benchmark better reflects the complexity and diversity of degradations encountered in real-world scenarios, rather than artifacts introduced by purely synthetic construction.

In total, the benchmark contains 464 non-reference degraded images for testing. To ensure a fair and consistent evaluation protocol, we adopt a fixed enhancement instruction for all samples. This design minimizes the influence of instruction variation and allows the evaluation to focus more directly on a model's restoration capability and its ability to preserve image consistency.

The collected images cover a variety of common real-world degradation scenarios, including complex and mixed degradations that are often challenging for restoration models. As a result, the benchmark provides a practical and demanding testbed for assessing real-world restoration performance. More details about the benchmark construction and data statistics are provided in Appendix~\ref{sec:benchmark}.

\begin{table*}[thbp]
\caption{Quantitative comparison on the Rain Removal, Deblurring, Low-light Enhancement, Haze Removal, and Reflection Removal tasks. 
We compared state-of-the-art (SOTA) image editing models. For each task, we reported LPS ($\downarrow$), RS ($\uparrow$), and FS ($\uparrow$). 
The best result is marked in \textbf{bold}, and \underline{underline} indicates the second-best result. The \colorbox{yellow!20}{best} and \colorbox{blue!20}{second-best} open-source results are highlighted with yellow and blue backgrounds, respectively.
}
\renewcommand{\arraystretch}{1.1}
\resizebox{\textwidth}{!}{
\begin{tabular}{l c | ccc | ccc | ccc | ccc | ccc}
\toprule
\multirow{2}{*}{Method} & \multirow{2}{*}{Open-source}
& \multicolumn{3}{c|}{Rain Removal}
& \multicolumn{3}{c|}{Deblurring}
& \multicolumn{3}{c|}{Low-light Enhancement}
& \multicolumn{3}{c|}{Haze Removal}
& \multicolumn{3}{c}{Reflection Removal}
\\
\cmidrule(lr){3-5}\cmidrule(lr){6-8}\cmidrule(lr){9-11}\cmidrule(lr){12-14}\cmidrule(lr){15-17}
& & LPS$\downarrow$ & RS$\uparrow$ & FS$\uparrow$ & LPS$\downarrow$ & RS$\uparrow$ & FS$\uparrow$ & LPS$\downarrow$ & RS$\uparrow$ & FS$\uparrow$ & LPS$\downarrow$ & RS$\uparrow$ & FS$\uparrow$ & LPS$\downarrow$ & RS$\uparrow$ & FS$\uparrow$ \\
\midrule
\multicolumn{17}{l}{\textit{Image Editing Methods}} \\
\midrule
Nano Banana Pro~\cite{team2023gemini} & No & 0.429 & \underline{2.063} & \textbf{0.236} & 0.326 & 1.068 & 0.144 & 0.467 & 0.720 & 0.077 & 0.492 & \textbf{1.920} & \textbf{0.195} & 0.358 & 1.368 & 0.176 \\
GPT\textendash Image\textendash 1.5~\cite{gpt4o20250325} & No & 0.535 & \textbf{2.120} & \underline{0.197} & 0.532 & \underline{1.667} & \underline{0.156} & 0.523 & \underline{1.048} & \underline{0.100} & 0.558 & \underline{1.840} & \underline{0.163} & 0.468 & \textbf{2.320} & \textbf{0.247} \\
Seedream 4.5~\cite{seedream2025seedream} & No & 0.438 & 1.500 & 0.169 & 0.254 & 0.255 & 0.038 & 0.423 & 0.600 & 0.069 & 0.418 & 1.140 & 0.133 & 0.291 & 1.156 & 0.164 \\
LongCat\textendash Image\textendash Edit~\cite{LongCat-Image} & Yes & 0.381 & \cellcolor{blue!20}1.302 & \cellcolor{blue!20}0.161 & 0.200 & 0.000 & 0.000 & 0.158 & 0.120 & 0.020 & 0.236 & 0.000 & 0.000 & 0.254 & 1.060 & 0.158 \\
Qwen\textendash Image\textendash Edit\textendash 2511~\cite{wu2025qwenimagetechnicalreport} & Yes & 0.435 & \cellcolor{yellow!20}1.736 & \cellcolor{yellow!20}0.196 & \cellcolor{blue!20}\underline{0.170} & 0.240 & 0.040 & \cellcolor{blue!20}\underline{0.122} & 0.080 & 0.014 & 0.337 & 0.060 & 0.008 & 0.333 & \cellcolor{yellow!20}\underline{1.820} & \cellcolor{yellow!20}\underline{0.243} \\
FLUX.1-Kontext-dev~\cite{labs2025flux} & Yes & \cellcolor{yellow!20}\textbf{0.244} & 0.673 & 0.102 & \cellcolor{yellow!20}\textbf{0.090} & 0.104 & 0.019 & \cellcolor{yellow!20}\textbf{0.108} & 0.160 & 0.029 & \cellcolor{yellow!20}\textbf{0.058} & 0.020 & 0.004 & \cellcolor{yellow!20}\textbf{0.048} & 0.127 & 0.024 \\
Step1X\textendash Edit~\cite{liu2025step1x} & Yes & \cellcolor{blue!20}\underline{0.282} & 0.019 & 0.003 & 0.321 & \cellcolor{blue!20}0.906 & \cellcolor{blue!20}0.123 & 0.306 & \cellcolor{blue!20}0.340 & \cellcolor{blue!20}0.047 & \cellcolor{blue!20}\underline{0.194} & \cellcolor{blue!20}0.190 & \cellcolor{blue!20}0.031 & \cellcolor{blue!20}\underline{0.247} & 0.080 & 0.012 \\
\midrule
\textbf{RealRestorer (ours)} & Yes & 0.371 & 1.076 & 0.135 & 0.582 & \cellcolor{yellow!20}\textbf{1.900} & \cellcolor{yellow!20}\textbf{0.159} & 0.597 & \cellcolor{yellow!20}\textbf{1.360} & \cellcolor{yellow!20}\textbf{0.110} & 0.339 & \cellcolor{yellow!20}0.680 & \cellcolor{yellow!20}0.090 & 0.290 & \cellcolor{blue!20}1.620 & \cellcolor{blue!20}0.230 \\
\bottomrule
\end{tabular}}
\label{tab:table1}
\end{table*}

\subsection{Experimental Results on RealIR-Bench}

% \subsubsection{Evaluation on \textbf{\Bench}}

Based on RealIR-Bench, we evaluate a diverse set of large image editing models' ability in image restoration towards the real world, covering state-of-the-art closed-source systems such as GPT-Image-1.5~\cite{gpt4o20250325}, Nano Banana Pro~\cite{team2023gemini}, Seeddream 4.5~\cite{seedream2025seedream}, as well as strong open models including Qwen-Image-Edit-2511~\cite{wu2025qwenimagetechnicalreport}, FLUX.1-Kontext-dev~\cite{labs2025flux}, LongCat-Image-Edit~\cite{LongCat-Image} and Step1X-Edit~\cite{liu2025step1x}. We provide nine major degradation tasks for evaluation: deblurring, rain removal, denoise, low-light enhancement, moiré patterns removal, haze removal, compression restoration, reflection removal, and deflare, with task-specific English instructions for each model to remove the corresponding degradation.

\begin{table*}[hbtp]
\centering
\caption{Quantitative comparison on the Deflare, Demoir\'e, Denoise, and Compression-restoration tasks. 
The average results of all 9 tasks are reported in the last column.
We compared state-of-the-art (SOTA) image editing models.
For each task, we reported LPIPS ($\downarrow$), RS ($\uparrow$), and FS ($\uparrow$). 
The best result is marked in \textbf{bold}, and \underline{underline} indicates the second-best result. The \colorbox{yellow!20}{best} and \colorbox{blue!20}{second-best} open-source results are highlighted with yellow and blue backgrounds, respectively.
}
\renewcommand{\arraystretch}{1.1}
\resizebox{\textwidth}{!}{
\begin{tabular}{l c | ccc | ccc | ccc | ccc | ccc}
\toprule
\multirow{2}{*}{Method} & \multirow{2}{*}{Open-source}
& \multicolumn{3}{c|}{Deflare}
& \multicolumn{3}{c|}{Moir\'e Patterns Removal}
& \multicolumn{3}{c|}{Denoise}
& \multicolumn{3}{c|}{Compression Restoration}
& \multicolumn{3}{c}{Avg Total (9)}
\\
\cmidrule(lr){3-5}\cmidrule(lr){6-8}\cmidrule(lr){9-11}\cmidrule(lr){12-14}\cmidrule(lr){15-17}
& & LPS$\downarrow$ & RS$\uparrow$ & FS$\uparrow$ & LPS$\downarrow$ & RS$\uparrow$ & FS$\uparrow$ & LPS$\downarrow$ & RS$\uparrow$ & FS$\uparrow$ & LPS$\downarrow$ & RS$\uparrow$ & FS$\uparrow$ & LPS$\downarrow$ & RS$\uparrow$ & FS$\uparrow$ \\
\midrule
\multicolumn{17}{l}{\textit{Image Editing Methods}} \\
\midrule
Nano Banana Pro~\cite{team2023gemini} & No & 0.214 & 1.222 & 0.192 & 0.562 & 1.560 & 0.137 & \underline{0.386} & 0.712 & 0.087 & 0.483 & 1.122 & \textbf{0.116} & 0.413 & 1.306 & \textbf{0.153} \\
GPT\textendash Image\textendash 1.5~\cite{gpt4o20250325} & No & 0.336 & 1.415 & 0.188 & 0.646 & \underline{1.633} & 0.116 & 0.496 & \textbf{0.993} & \textbf{0.100} & 0.633 & \textbf{1.167} & 0.086 & 0.525 & \textbf{1.578} & \underline{0.150} \\
Seedream 4.5~\cite{seedream2025seedream40nextgenerationmultimodal} & No & 0.225 & 1.104 & 0.171 & 0.548 & 1.600 & \textbf{0.145} & 0.387 & 0.770 & \underline{0.094} & 0.529 & \underline{1.136} & \underline{0.107} & 0.390 & 1.029 & 0.125 \\
LongCat\textendash Image\textendash Edit~\cite{LongCat-Image} & Yes & 0.241 & \cellcolor{yellow!20}\textbf{1.717} & \cellcolor{yellow!20}\textbf{0.261} & \cellcolor{blue!20}\underline{0.420} & 1.200 & \cellcolor{blue!20}0.139 & \cellcolor{yellow!20}\textbf{0.350} & 0.471 & 0.061 & \cellcolor{yellow!20}\textbf{0.188} & 0.083 & 0.014 & \cellcolor{blue!20}\underline{0.270} & 0.661 & 0.097 \\
Qwen\textendash Image\textendash Edit\textendash 2511~\cite{wu2025qwenimagetechnicalreport} & Yes & 0.222 & \cellcolor{blue!20}\underline{1.660} & \cellcolor{blue!20}\underline{0.258} & 0.595 & \cellcolor{yellow!20}\textbf{1.660} & 0.135 & 0.429 & \cellcolor{blue!20}0.824 & \cellcolor{yellow!20}0.094 & \cellcolor{blue!20}\underline{0.242} & 0.300 & 0.046 & 0.320 & \cellcolor{blue!20}0.931 & \cellcolor{blue!20}0.127 \\
FLUX.1-Kontext-dev~\cite{labs2025flux} & Yes & \cellcolor{yellow!20}\textbf{0.064} & 0.264 & 0.049 & \cellcolor{yellow!20}\textbf{0.348} & 0.540 & 0.070 & 0.429 & 0.628 & 0.072 & 0.429 & \cellcolor{blue!20}0.628 & \cellcolor{blue!20}0.072 & \cellcolor{yellow!20}\textbf{0.202} & 0.349 & 0.056 \\
Step1X\textendash Edit~\cite{liu2025step1x} & Yes & \cellcolor{blue!20}\underline{0.173} & 0.000 & 0.000 & 0.654 & 0.410 & 0.028 & \cellcolor{blue!20}0.409 & 0.098 & 0.012 & 0.344 & 0.383 & 0.050 & 0.325 & 0.270 & 0.036 \\
\midrule
\textbf{RealRestorer (ours)} & Yes & 0.239 & 1.623 & 0.247 & 0.563 & \cellcolor{blue!20}1.620 & \cellcolor{yellow!20}\underline{0.142} & 0.478 & \cellcolor{yellow!20}\underline{0.863} & \cellcolor{blue!20}0.090 & 0.547 & \cellcolor{yellow!20}1.067 & \cellcolor{yellow!20}0.097 & 0.445 & \cellcolor{yellow!20}\underline{1.312} & \cellcolor{yellow!20}0.146 \\
\bottomrule
\end{tabular}}
\label{tab:cmp2}
\end{table*}

\begin{table*}[thbp]
\centering
\caption{Quantitative comparison on the FoundIR dataset across various real-world degradations. We report PSNR ($\uparrow$) and SSIM ($\uparrow$). The best results are highlighted in \textbf{bold}, and the second-best results are \underline{underlined}.}
\label{tab:foundir_comparison}
\resizebox{\textwidth}{!}{
\begin{tabular}{l | cc | cc | cc | cc | cc | cc | cc | cc}
\toprule
\multirow{2}{*}{Method}
& \multicolumn{2}{c|}{\textbf{Blur}} 
& \multicolumn{2}{c|}{\textbf{Rain}} 
& \multicolumn{2}{c|}{\textbf{Raindrops}} 
& \multicolumn{2}{c|}{\textbf{Noise}} 
& \multicolumn{2}{c|}{\textbf{Low-light}} 
& \multicolumn{2}{c|}{\textbf{Haze}} 
& \multicolumn{2}{c|}{\textbf{Compression}} 
& \multicolumn{2}{c}{\textbf{Average}} \\
\cmidrule(lr){2-3} \cmidrule(lr){4-5} \cmidrule(lr){6-7} \cmidrule(lr){8-9} \cmidrule(lr){10-11} \cmidrule(lr){12-13} \cmidrule(lr){14-15} \cmidrule(lr){16-17}
 & PSNR$\uparrow$ & SSIM$\uparrow$ & PSNR$\uparrow$ & SSIM$\uparrow$ & PSNR$\uparrow$ & SSIM$\uparrow$ & PSNR$\uparrow$ & SSIM$\uparrow$ & PSNR$\uparrow$ & SSIM$\uparrow$ & PSNR$\uparrow$ & SSIM$\uparrow$ & PSNR$\uparrow$ & SSIM$\uparrow$ & PSNR$\uparrow$ & SSIM$\uparrow$ \\
\midrule
Nano Banana Pro~\cite{team2023gemini} & \textbf{20.16} & \textbf{0.68} & \underline{21.55} & \underline{0.65} & \underline{23.29} & \underline{0.66} & 24.78 & 0.89 & \underline{15.73} & \underline{0.73} & \underline{17.63} & \underline{0.50} & 19.48 & 0.58 & \underline{20.37} & \underline{0.67} \\
GPT\textendash Image\textendash 1.5~\cite{gpt4o20250325} & 13.25 & 0.41 & 12.52 & 0.34 & 13.89 & 0.31 & 14.15 & 0.68 & 11.00 & 0.55 & 12.39 & 0.28 & 13.06 & 0.39 & 12.89 & 0.42 \\
Seedream 4.5~\cite{seedream2025seedream} & 17.94 & 0.61 & 17.61 & 0.57 & 18.01 & 0.58 & 15.57 & 0.78 & 13.64 & 0.68 & 15.65 & 0.44 & 14.29 & 0.54 & 16.10 & 0.60 \\
LongCat\textendash Image\textendash Edit~\cite{LongCat-Image} & 18.53 & \underline{0.65} & 17.09 & 0.52 & 18.22 & 0.48 & 19.77 & 0.81 & 15.24 & 0.70 & 10.52 & 0.36 & 15.64 & 0.47 & 16.43 & 0.57 \\
Qwen\textendash Image\textendash Edit\textendash 2511~\cite{wu2025qwenimagetechnicalreport} & 15.60 & 0.55 & 14.68 & 0.46 & 15.19 & 0.42 & 19.73 & 0.81 & 7.13 & 0.17 & 7.20 & 0.34 & 18.29 & 0.61 & 13.98 & 0.48 \\
FLUX.1-Kontext-dev~\cite{labs2025flux} & 10.73 & 0.41 & 10.48 & 0.35 & 11.25 & 0.31 & 15.10 & 0.69 & 10.34 & 0.53 & 10.74 & 0.29 & 10.74 & 0.34 & 11.34 & 0.42 \\
Step1X\textendash Edit~\cite{liu2025step1x} & 16.38 & 0.63 & 19.91 & 0.63 & 19.42 & 0.58 & \underline{27.18} & \textbf{0.91} & 15.72 & 0.70 & 14.20 & 0.45 & \textbf{22.81} & \textbf{0.75} & 19.37 & 0.66 \\
\midrule
\textbf{RealRestorer (ours)} & \underline{18.99} & 0.59 & \textbf{23.72} & \textbf{0.71} & \textbf{23.64} & \textbf{0.71} & \textbf{28.15} & \underline{0.90} & \textbf{17.59} & \textbf{0.77} & \textbf{17.66} & \textbf{0.56} & \underline{20.40} & \underline{0.66} & \textbf{21.45} & \textbf{0.70} \\
\bottomrule
\end{tabular}}
\end{table*}

% We compare our method with large-scale image editing models, such as Seedream 4.5, Nano Banana Pro, GPT-Image-1.5, Step1X-Edit, FLUX.1-Kontext-dev, Qwen-Image-Edit-2511, and LongCat-Image-Edit.
% Our approach consistently produces cleaner structures, fewer artifacts, and better perceptual quality across diverse degradations. These results demonstrate the strong generalization ability of our method under both seen and unseen degradation scenarios.

Unlike full-reference metrics such as PSNR and SSIM~\cite{wang2004image}, which require paired clean reference images for evaluation, RealIR-Bench is built entirely from non-reference images collected from diverse real-world scenarios. In these cases, obtaining perfectly aligned clean targets is infeasible, making conventional full-reference evaluation protocols unsuitable. Therefore, instead of relying on pixel-wise fidelity measures, we adopt a non-reference evaluation framework to assess how well image editing models handle real-world degradations, with particular emphasis on both degradation removal capability and consistency preservation.

To characterize both restoration effectiveness and the trade-off with content fidelity, we report two metrics: Restoration Score (RS), LPIPS (LPS)~\cite{zhang2018unreasonable}. We convert the LPIPS distance into a perceptual similarity score so that higher values indicate better perceptual consistency. After normalizing both RS and LPS to the same scale, the Final Score (FS) is defined as:
\begin{equation}
FS = 0.2 \, (1 - LPS) \, RS
\end{equation}
FS jointly reflects restoration improvement and content preservation, and poor performance in either aspect will directly lead to a lower overall score.

Inspired by non-reference evaluation methods such as VIEScore~\cite{ku2024viescore}, we leverage VLMs to assess  Restoration Score. Specifically, we employ Qwen3-VL-8B-Instruct~\cite{qwen3technicalreport} to rate the degradation severity of both degraded images and restored images on a scale from 0 to 5, where 5 indicates no visible degradation, and 0 corresponds to the most severe degradation.
The Restoration Score (RS) is defined as the improvement in degradation level after restoration. In other words, it is computed as the difference between the degradation score of the restored image and that of the degraded image. A higher RS indicates greater perceived restoration improvement according to the VLM evaluator.

For consistency evaluation, we aim to measure the model’s ability to preserve the original scene structure, semantic content, and fine-grained details throughout the restoration process. To this end, we employ LPIPS as the evaluation metric to measure the perceptual similarity between the restored images and the degraded inputs. Unlike traditional pixel-level metrics, LPIPS is more sensitive to perceptually relevant discrepancies, including structural deviations and semantic inconsistencies, making it particularly suitable for assessing content preservation before and after restoration.

Table~\ref{tab:table1} and Table~\ref{tab:cmp2} demonstrate the strong restoration capability of RealRestorer. It consistently outperforms existing open-source image editing models and achieves performance comparable to leading closed-source systems. Across all nine tasks, RealRestorer achieves the best performance on deblurring and low-light enhancement and ranks second on moiré pattern removal. Among open-source models, it ranks first on five tasks and second on two, and remains highly competitive on the remaining tasks. Overall, it ranks \textbf{first among open-source models} and third overall, narrowing the gap with Nano Banana Pro (first place) to only 0.007 points and surpassing Qwen-Image-Edit-2511 (the second-best open-source model) by 0.019 points. These results indicate that RealRestorer not only effectively removes real-world degradations but also maintains high consistency and fidelity in the restored outputs. As an open-source model, RealRestorer significantly narrows the performance gap between open-source and closed-source systems, while exhibiting strong generalization ability across diverse real-world scenarios. Figure~\ref{fig:comp} presents qualitative comparisons, further demonstrating that RealRestorer produces visually cleaner and more consistent restoration results compared to other state-of-the-art image editing methods. On real-world degraded images from RealIR-Bench, our model shows strong performance across diverse scenarios. In particular, when handling complex and irregular real-world degradations such as blur and flare, RealRestorer remains 
highly competitive with leading closed-source models, achieving comparable visual quality and structural fidelity.

% To evaluate the generalization ability of RealRestorer, we conduct zero-shot task generalization experiments in real-world scenarios, including snow removal, old photo restoration, shadow removal, and watermark removal. As shown in Figure~\ref{fig:genelazation}, RealRestorer demonstrates strong adaptability when encountering previously unseen tasks. Despite not being explicitly trained for these specific degradations, the model is able to leverage its learned restoration knowledge to produce visually plausible and consistent results.

% We attribute this strong generalization capability to three key factors. First, the diversity and richness of the real-world degradation data expose the model to a wide range of degradation patterns during training, enabling it to learn more comprehensive degradation representations. Second, the underlying image editing backbone provides powerful prior knowledge acquired from large-scale pretraining, which facilitates cross-task transfer. Finally, by freezing the text encoder during training, we preserve its semantic understanding and alignment ability, allowing the model to recognize and respond to degradation types that were not explicitly observed during training.

% Together, these factors enable RealRestorer to generalize beyond seen degradations and behave as a more versatile restoration model in practical scenarios.

\subsection{Extra Benchmark Evaluation and Zero-shot Generalization}

To further evaluate restoration performance on a traditional all-in-one benchmark, we additionally evaluate the same set of image editing models on the FoundIR test set~\cite{li2025foundir}. FoundIR contains 20 real-world degradation settings with paired clean references, including 7 isolated degradations (blur, rain, noise, low-light, raindrops, haze, and compression artifacts) and 13 coupled degradation combinations. We report results on the 7 isolated degradation subsets, which also overlap with RealIR-Bench, resulting in a total of 750 paired image pairs with an average resolution of 2514 × 1516. For the editing prompt, we use the same prompt set as RealIR-Bench.

% \begin{figure*}[htbp]
%     \centering
%     % \setlength{\abovecaptionskip}{2pt}
%     % \vspace{-2mm}
%     \includegraphics[width=0.98\linewidth]{img/genelazation.pdf}
%     \caption{\small \textbf{Zero-shot generalization results of RealRestorer on unseen real-world degradation tasks, including snow removal and old photo restoration.}}
%     \label{fig:genelazation}
% \end{figure*}

% \begin{figure}[htbp]
%     \centering
%         \centering
%         \includegraphics[width=\linewidth]{img/genelazation.pdf}
%         \caption{\small \textbf{Zero-shot generalization results of RealRestorer on unseen real-world degradation tasks, including snow removal and old photo restoration.}}
%         \label{fig:genelazation}
% \end{figure}

Table~\ref{tab:foundir_comparison} shows that RealRestorer achieves strong restoration performance on these 7 tasks, obtaining the best PSNR and SSIM on 5 out of 7 degradations. Notably, all image editing models tend to achieve relatively low reference-based metrics, which is consistent with the generative nature of these models that may introduce perceptually plausible yet non-identical details. Benefiting from high-quality synthetic degradation data, RealRestorer achieves a better trade-off while improving content consistency.
We further evaluate the generalization ability of RealRestorer via zero-shot experiments on real-world restoration scenarios, including snow removal and old photo restoration. RealRestorer also generalizes well to unseen restoration tasks. Although it is only fine-tuned on a limited set of degradation types, it can still handle other unseen tasks by benefiting from the restoration priors learned during training, while retaining part of the original model’s general image editing capability. More qualitative results, evaluations on additional public benchmarks, and detailed comparisons are provided in Appendix~\ref{sec:extra_experiment}, together with further visualizations and analysis.

\subsection{Ablation and User Studies}

\begin{figure}[htbp]
    \centering
        \centering
        \includegraphics[width=\linewidth]{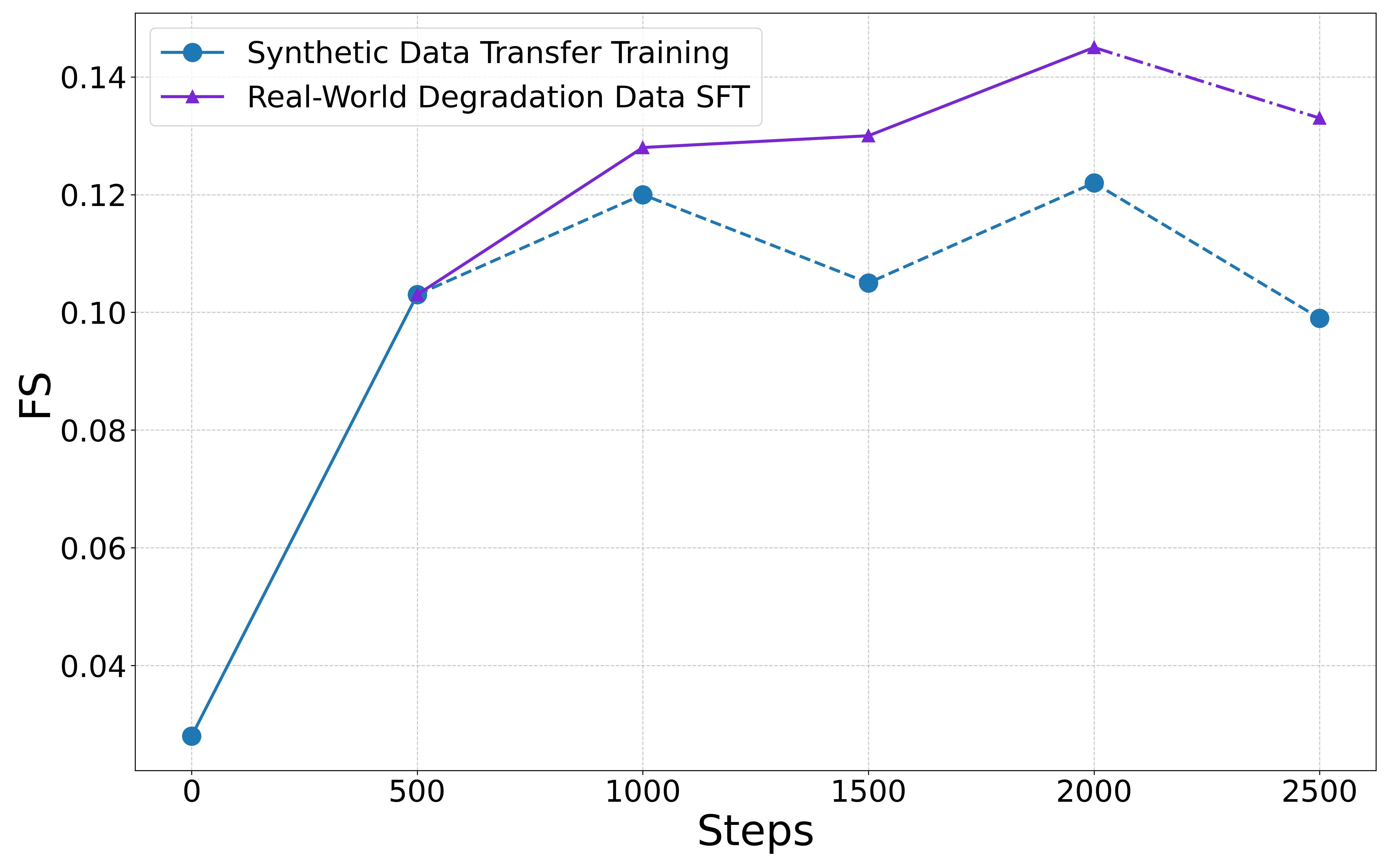}
        \captionof{figure}{ \small Model performance with varying training steps and training data on RealIR-Bench. The blue line shows transfer training on synthetic degradation data, where the model gradually acquires basic restoration capability. The blue dashed line indicates performance degradation after prolonged training due to the limited diversity of synthetic data. The purple line represents supervised fine-tuning with real-world degradation data, which rapidly improves performance and generalization. The purple dashed segment indicates the onset of overfitting after around 2.5K steps.}
        \label{fig:ablation}
\end{figure}
\vspace{-3mm}
We conduct an ablation study on the training data and training stages to examine the necessity of the proposed two-stage training strategy. Specifically, we first train the model on the Synthetic Degradation Data (about 1M samples). As shown in Figure~\ref{fig:ablation}, the model acquires basic restoration capability and reaches a peak FS of 0.122 during the first stage, but still lacks sufficient generalization ability and fails on some rare cases. Moreover, its performance drops significantly after 2.5K steps, which we attribute to the limited diversity of the synthetic data. We further investigate the impact of the Real-World Degradation Data (about 100K samples) in the second stage. After entering this stage, the model quickly surpasses the peak score achieved in the transfer training stage and continues to improve its generalization ability, eventually achieving strong performance at around  2.5K steps. However, beyond this point, the model begins to overfit the Real-World Degradation Data, which motivates us to adopt early stopping. Overall, the two-stage training strategy, together with the combination of synthetic and real-world data, leads to a final model with strong restoration performance and better consistency preservation.

Furthermore, we conduct an ablation study on the Progressively-Mixed training strategy. Without this component, the final FS score decreases by 0.004 points under the same training configuration, confirming its effectiveness. And from a qualitative perspective, the Progressive-Mixed strategy also leads to better preservation of structural consistency and content fidelity, resulting in more visually stable and coherent restoration results. Additional ablation results and analyses are provided in the supplementary materials.

% \begin{figure*}[htbp]
%     \centering
%     % \setlength{\abovecaptionskip}{2pt}
%     % \vspace{-2mm}
%     \includegraphics[width=0.98\linewidth]{img/genelazation.pdf}
%     \caption{\small \textbf{Zero-shot generalization results of RealRestorer on unseen real-world degradation tasks, including snow removal and old photo restoration.}}
%     \label{fig:genelazation}
% \end{figure*}

We conduct a user study to evaluate both the reliability of the proposed RealIR-Bench metrics and the perceptual performance of our model from a human perspective. Specifically, we recruit 32 participants to rank 3,200 groups of generated images produced by five high-performing models according to two criteria: restoration quality and content consistency. Specifically, Nano Banana Pro achieves the highest first-ranking rate of 32.02\%, followed by GPT-Image-1.5 with 23.83\%, while our method attains 21.54\%. This trend is consistent with the average overall scores reported in Table~\ref{tab:cmp2}. Moreover, we perform a statistical analysis of the proposed metrics and observe a moderate alignment with human judgments across all evaluation measures ($p<0.01$). Further details on the study design, ranking protocol, and statistical analysis are provided in the Apeendix~\ref{sec:ablation}.

% \section{\ours : Controllable and ID-Consistent Generation}\label{sec:method}

\section{Limitations and Discussion}
Although RealRestorer demonstrates strong generalization across both seen and unseen restoration tasks, we still observe several limitations. First, since the base image editing model relies on a 28-step denoising process, its computational cost remains substantially higher than that of smaller-scale models, which is a common limitation of large-scale image editing models. Second, in cases with strong semantic and physical ambiguity, such as mirror selfies, the model may fail to distinguish true scene content from undesired reflections, a challenge that is also common in other image editing methods. Third, RealRestorer still struggles with extremely severe degradations where reliable pixel evidence is largely missing, and may fail to preserve physically consistent structures such as water reflections.

\section{Conclusion}
In this paper, we introduce \textbf{RealRestorer}, a robust open-source image editing model for complex real-world image restoration. To reduce the synthetic-to-real domain gap, we propose a comprehensive data generation pipeline and a two-stage progressively mixed training strategy that combines synthetic and real-to-clean pairs. We further present \textbf{RealIR-Bench}, a non-reference benchmark with authentic degraded images and a VLM-based evaluation framework for real-world restoration. Extensive experiments on many evaluation sets demonstrate that \textbf{RealRestorer} achieves open-source state-of-the-art performance across nine restoration tasks, with results highly comparable to leading closed-source commercial systems, and exhibits strong zero-shot generalization to unseen degradations. We will release our model, data synthesis pipeline, and benchmark to support future research in real-world image restoration.

% \section{Experiments}\label{sec:evaluation}
% \input{sections/6.evaluation}

% % Discussion and Conclusion. Section inside
% \newpage % manuel balance
% \input{sections/7.conclusion}

\clearpage
\newpage
% \columnbreak
{
\bibliography{main}
\bibliographystyle{plain}
}


\clearpage
\newpage

\clearpage
\appendix
\noindent
\textbf{\LARGE Appendix}
\vspace{5ex}
\setcounter{section}{0}
\setcounter{subsection}{0}

\renewcommand{\thesection}{\Alph{section}}
\renewcommand{\thesubsection}{\thesection.\arabic{subsection}}

\section{Data Construction Details}
\label{sec:Data}
\subsection{Synthetic Degradation Data}
For the Synthetic Degradation Data, we collect the clean image from the internet and synthesize the nine major degradation patterns: blur, rain, noise, low-light, moiré patterns, haze, compression artifacts, reflection, and flare. We will release the pipeline.

\subsection{Real-World Degradation Data}
For the real-world degradation data, we collect clean images from high-quality open-source image websites, including Pexels and Pinterest, covering six types of degradation: blur, rain, low light, haze, reflection, and flare. These degradation types often exhibit a substantial gap between real-world degradations and synthesized patterns.

% \begin{table}[tbhp]
% \centering
% \caption{Semantic prompts used for CLIP-based degradation filtering.}
% \label{tab:clip_prompts}
% \renewcommand{\arraystretch}{1.2}
% \begin{tabular}{p{3cm} p{9cm}}
% \toprule
% \textbf{Degradation Type} & \textbf{CLIP Text Prompt} \\
% \midrule
% Flare & a photo with lens flare, bright streaks of light \\
% Haze & a hazy photo, foggy atmosphere, low contrast \\
% Rain & a rainy photo with rain streaks or raindrops \\
% Low-light & a dark photo, underexposed, low illumination \\
% Blur & a blurry photo with motion blur or out-of-focus regions \\
% Reflection & a photo with glass reflection or mirror-like reflection artifacts \\
% \bottomrule
% \end{tabular}
% \end{table}

\begin{table}[tbhp] % 去掉星号，恢复单栏排版
\centering
\caption{Semantic prompts used for CLIP-based degradation filtering.}
\label{tab:clip_prompts}
\renewcommand{\arraystretch}{1.2}
% 自动将表格等比压缩到当前单栏的宽度 (\columnwidth)
\resizebox{\columnwidth}{!}{%
\begin{tabular}{@{}ll@{}} % @{} 作用是去掉表格最左侧和最右侧的内边距，挤出更多空间
\toprule
\textbf{Degradation Type} & \textbf{CLIP Text Prompt} \\
\midrule
Flare      & a photo with lens flare, bright streaks of light \\
Haze       & a hazy photo, foggy atmosphere, low contrast \\
Rain       & a rainy photo with rain streaks or raindrops \\
Low-light  & a dark photo, underexposed, low illumination \\
Blur       & a blurry photo with motion blur or out-of-focus regions \\
Reflection & a photo with glass or mirror-like reflection artifacts \\
\bottomrule
\end{tabular}%
}
\end{table}

To construct a high-quality real-world degradation dataset, we first employ the CLIP model~\cite{radford2021learning} to filter images based on degradation-related semantic cues, as shown in Table~\ref{tab:clip_prompts}. Second, we apply a watermark detection filter~\cite{minderer2023scaling} together with Qwen3-VL-8B-Instruct~\cite{qwen3technicalreport} to remove watermarked images and images with insufficient degradation, thereby retaining samples suitable for obtaining paired clean data. After selecting appropriate editing models to generate a large amount of raw paired data, additional filtering is required to remove failure cases. Specifically, we use Qwen3-VL-8B-Instruct to estimate the degradation scores of both the clean and degraded images, and then filter pairs with inconsistent or insufficient score differences, while a skeleton-shift-based method~\cite{cheng2020skeleton} is adopted to remove pixel pairs with alignment errors. Finally, after strictly filtering the raw dataset, we further performed human curation on the remaining subset to construct the final dataset. 
Three trained human experts participated in the annotation and verification process.

\subsection{Training Dataset statistics}
\begin{figure*}[thbp]
    \centering
    \setlength{\abovecaptionskip}{2pt}
    \includegraphics[width=0.98\linewidth]{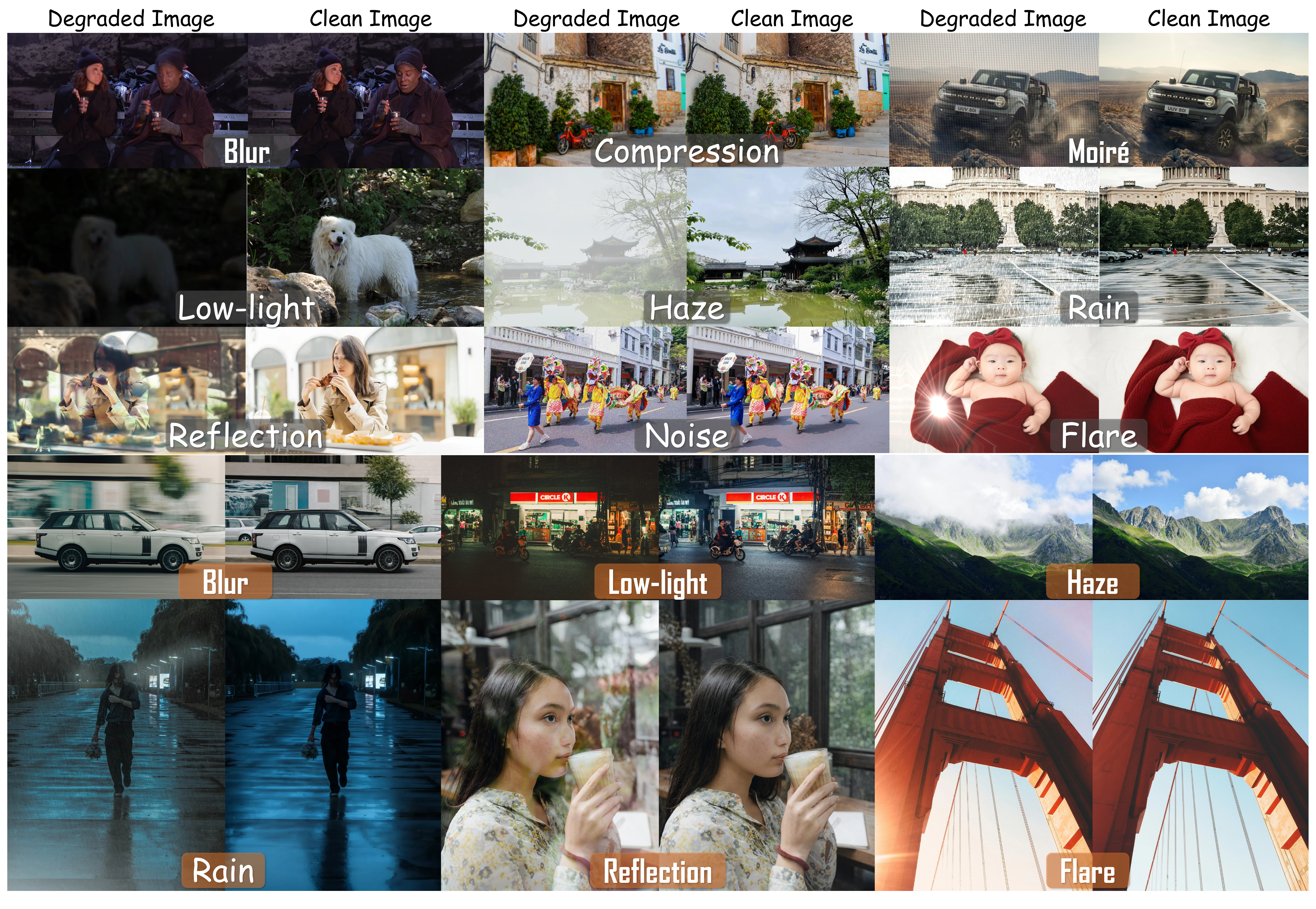}
    \vspace{0.5ex}
    \caption{\small Examples from our training dataset containing both synthetic and real-world degradation pairs. The upper rows with gray labels show synthesized degradations generated by our pipeline, while the bottom rows highlighted with orange labels correspond to real-world degraded images paired with clean references.}
    \vspace{-3ex}
    \label{fig:dataconstruct1}
\end{figure*}

Table~\ref{tab:data} summarizes the statistics of the two components of our training data across different degradation types. 
Visual examples of the data are presented in Figure~\ref{fig:dataconstruct1}.

% \begin{table}[h]
% \centering
% \caption{Statistics of our dataset across different degradation types. 
% The table reports the number of training image pairs from Synthetic Degradation Data and Real-World Degradation Data. 
% The total column indicates the combined number of samples for each degradation category.}
% \label{tab:data}
% \begin{tabular*}{\textwidth}{l@{\extracolsep{\fill}}rrr}
% \toprule
% \textbf{Degradation} & \textbf{Synthetic} & \textbf{Real} & \textbf{Total} \\
% \midrule
% Rain        & 84,968    & 43,415  & 128,383 \\
% Blur        & 1,014,229 & 13,458  & 1,027,687 \\
% Low-light   & 5,000     & 7,005   & 12,005 \\
% Hazy        & 103,971   & 8,147   & 112,118 \\
% Reflection  & 68,227    & 7,604   & 75,831 \\
% Flare       & 59,520    & 7,956   & 67,476 \\
% Moire       & 99,085    & 0       & 99,085 \\
% Noise       & 64,492    & 0       & 64,492 \\
% Compression & 68,000    & 0       & 68,000 \\
% \midrule
% \textbf{Total} & \textbf{1,567,492} & \textbf{87,585} & \textbf{1,655,077} \\
% \bottomrule
% \end{tabular*}
% \end{table}

\begin{table}[tbhp]
\centering
\caption{Statistics of our dataset across different degradation types. 
The table reports the number of training image pairs from Synthetic Degradation Data and Real-World Degradation Data. 
The total column indicates the combined number of samples for each degradation category.}
\label{tab:data}
% 注意这里将 \textwidth 改为了 \columnwidth
\begin{tabular*}{\columnwidth}{l@{\extracolsep{\fill}}rrr}
\toprule
\textbf{Degradation} & \textbf{Synthetic} & \textbf{Real} & \textbf{Total} \\
\midrule
Rain        & 84,968    & 43,415  & 128,383 \\
Blur        & 1,014,229 & 13,458  & 1,027,687 \\
Low-light   & 5,000     & 7,005   & 12,005 \\
Hazy        & 103,971   & 8,147   & 112,118 \\
Reflection  & 68,227    & 7,604   & 75,831 \\
Flare       & 59,520    & 7,956   & 67,476 \\
Moire       & 99,085    & 0       & 99,085 \\
Noise       & 64,492    & 0       & 64,492 \\
Compression & 68,000    & 0       & 68,000 \\
\midrule
\textbf{Total} & \textbf{1,567,492} & \textbf{87,585} & \textbf{1,655,077} \\
\bottomrule
\end{tabular*}
\end{table}

\section{Implementation Details}
\label{sec:train}

During training, we treat the DiT blocks as trainable components, while freezing both the VAE and text encoders. In the Transfer Training Stage, we train the model using the Synthetic Degradation Dataset, which covers nine degradation types. To balance the learning across tasks, we adopt an average sampling strategy over all nine degradation categories. In this stage, the bucket resolution is fixed at 1024 × 1024, and the global batch size is set to 16.

After approximately 500 training steps, the model begins to transfer knowledge from high-level editing capabilities to low-level restoration tasks. However, it still struggles to handle more complex degradations, often producing artifacts in the restored results. To address this limitation, we introduce a Supervised Fine-Tuning Stage. In this stage, we adopt a Progressively-Mixed training strategy, combining Real-World Degradation Data with a small portion of Synthetic Degradation Data. This strategy helps constrain the model toward the data manifold of real-world restoration tasks while retaining the robustness learned from synthetic degradations.

Additionally, we freeze the first one-fourth of the SingleStreamBlocks in the DiT architecture to stabilize training. The global batch size is increased to 32, and a cosine annealing learning rate schedule is applied, where the learning rate gradually decays to zero while maintaining the same initial learning rate as in the first stage. This stage lasts for 1.5K training steps, allowing the model to converge to a balanced and generalizable checkpoint. All experiments are conducted on NVIDIA H800 GPUs, and the entire training process takes approximately one day on 8 H800 GPUs. More detailed training hyperparameters are provided in Table~\ref{tab:hyperparameters}.

% table settings
% \begin{table}[t]
% \centering
% \caption{Training hyperparameters used in the two training stages.}
% \label{tab:hyperparameters}
% \begin{tabular}{lcc}
% \toprule
% \textbf{Hyperparameters} & \textbf{Transfer Training Stage} & \textbf{Supervised Fine-Tuning Stage} \\
% \midrule
% Learning rate & $1\times10^{-5}$ & $1\times10^{-5}\rightarrow0$ \\
% LR scheduler & Constant & Cosine \\
% Weight decay & 0.0 & 0.01 \\
% Gradient norm clip & 1.0 & 1.0 \\
% Optimizer & \multicolumn{2}{c}{AdamW ($\beta_1=0.9,\beta_2=0.95,\epsilon=1e^{-8}$)} \\
% Warm-up steps & 100 & 100 \\
% Frozen layers & None & First $1/4$ SingleStreamBlocks \\
% Training steps & 500 & 1500 \\
% Training samples & 1.5M & 80K \\
% Resolution & 1024 × 1024 & 1024 × 1024 \\
% Synthetic : Real data ratio & 1 : 0 & 2 : 8 \\
% \bottomrule
% \end{tabular}
% \end{table}
\begin{table*}[t] % 使用 table* 星号环境使其横跨双栏，[t]表示放在页顶
\centering
\caption{Training hyperparameters used in the two training stages.}
\label{tab:hyperparameters}

% 使用 tabular* 并指定 \textwidth，\extracolsep{\fill} 会自动均匀推开三列
\begin{tabular*}{\textwidth}{l@{\extracolsep{\fill}}cc}
\toprule
\textbf{Hyperparameters} & \textbf{Transfer Training Stage} & \textbf{Supervised Fine-Tuning Stage} \\
\midrule
Learning rate & $1\times10^{-5}$ & $1\times10^{-5}\rightarrow0$ \\
LR scheduler & Constant & Cosine \\
Weight decay & 0.0 & 0.01 \\
Gradient norm clip & 1.0 & 1.0 \\
Optimizer & \multicolumn{2}{c}{AdamW ($\beta_1=0.9, \beta_2=0.95, \epsilon=1e^{-8}$)} \\
Warm-up steps & 100 & 100 \\
Frozen layers & None & First $1/4$ SingleStreamBlocks \\
Training steps & 500 & 1500 \\
Training samples & 1.5M & 80K \\
Resolution & $1024 \times 1024$ & $1024 \times 1024$ \\
Synthetic : Real data ratio & 1 : 0 & 2 : 8 \\
\bottomrule
\end{tabular*}

\end{table*}

\section{RealIR-Bench and Metrics Details} \label{sec:benchmark}
\begin{figure*}[thbp]
    \centering
    \vspace{-18mm}
    \includegraphics[width=0.98\linewidth]{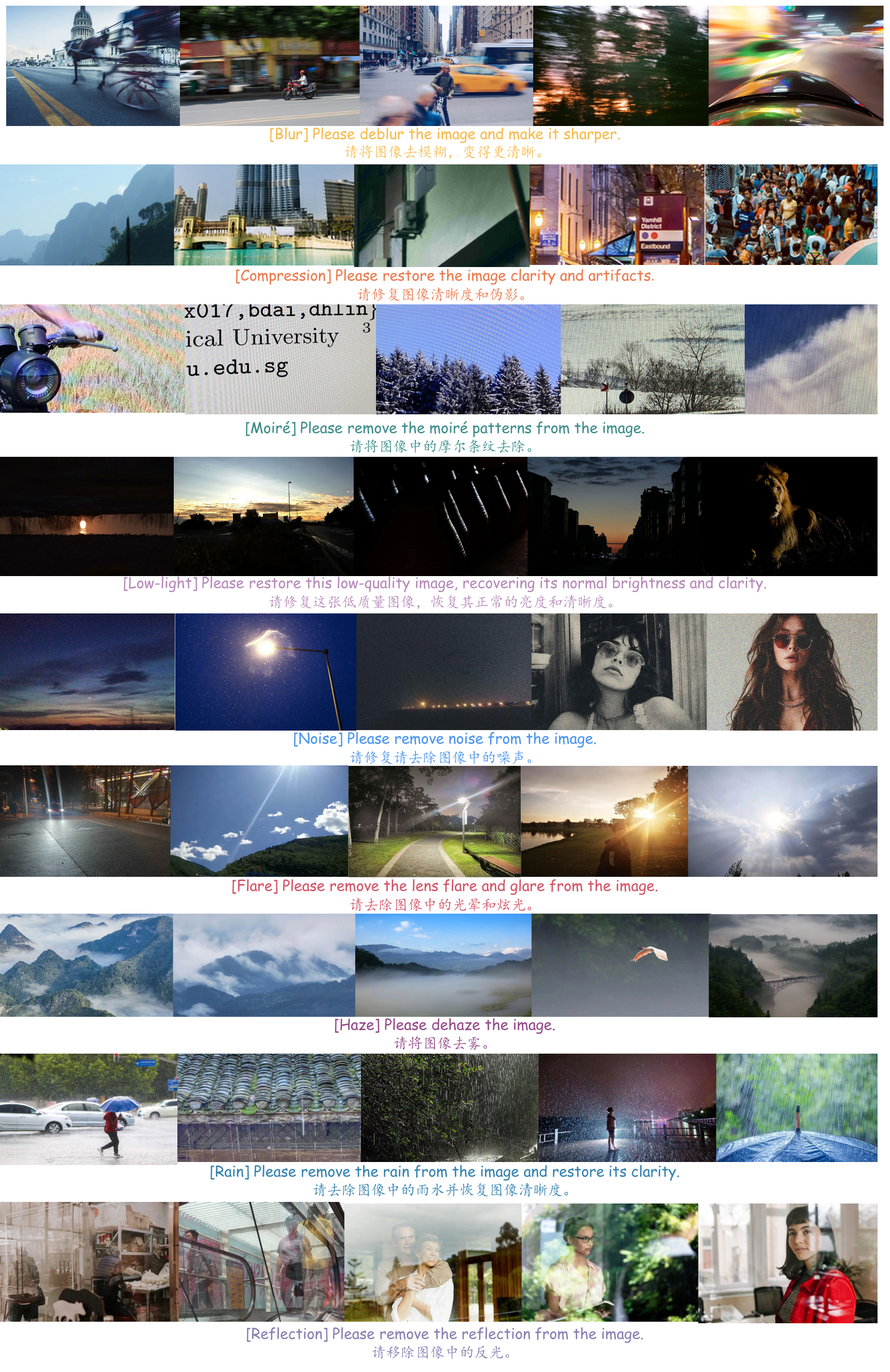}
    \caption{\small Examples from our RealIR-Bench. Each degradation category is evaluated using a fixed bilingual prompt.}
    \label{fig:benchmark_fig}
\end{figure*}

\textbf{RealIR-Bench} covers diverse real-world degradation scenarios, including blur, rain, noise, low-light, moiré patterns, haze, compression artifacts, reflection, and flare. Example cases from the benchmark are shown in Figure~\ref{fig:benchmark_fig}.

Specifically, we evaluate models using two complementary metrics: Restoration Score (RS), which reflects the perceptual restoration quality, and LPIPS~\cite{zhang2018unreasonable}, which measures perceptual similarity to assess consistency after restoration.

\subsection{Restoration Score}
The Restoration Score (RS) is designed to evaluate the ability of a model to remove degradations without explicitly considering content consistency. Inspired by VIEScore, we employ Qwen3-VL-8B-Instruct as a vision-language evaluator to assess the degradation severity of both degraded images and restored images. The Restoration Score (RS) is then defined as the improvement in the degradation level after restoration. The detailed system instruction for Qwen3-VL-8B-Instruct is shown in the Figure~\ref{fig:degradation_instruction}.

\section{More Qualitative Results and Benchmark Evaluation}
\label{sec:extra_experiment}
We present additional visualization results in the following pages to further demonstrate the strong restoration capability of our model compared with other image editing models on \textbf{RealIR-Bench} across nine degradation types.

Furthermore, to evaluate \textbf{RealRestorer} on public benchmarks for deflare, reflection removal, and demoiré, we conduct additional experiments on several widely used datasets. For deflare evaluation, we use the Flare-R subset from the Flare7K++ dataset~\cite{dai2023flare7kpp}, which contains 100 paired real-world flare images. The Flare7K++ dataset combines synthetic and real flare data and provides a comprehensive benchmark for nighttime flare removal tasks.

For moiré pattern removalevaluation, we adopt the UHDM test set~\cite{dai2022video}, which contains 500 paired real moiré images captured at ultra-high resolution. For reflection removal, we evaluate on the SIR²+ benchmark, which includes three subsets: SolidObjectDataset, PostcardDataset, and WildScene, containing 50, 50, and 101 paired images respectively. These datasets include real-world scenes with complex reflective patterns and are widely used for evaluating single-image reflection removal methods. The quantitative comparison results are presented in Table~\ref{tab:extra_comparison}. The results show that \textbf{RealRestorer} achieves the second-best performance in PSNR and the third-best performance in SSIM on average across the five evaluation datasets.

% 表二额外的补充
\begin{table*}[t]
\centering
\caption{
Quantitative comparison on extra public benchmarks for real-world image restoration tasks. 
We report PSNR ($\uparrow$) and SSIM ($\uparrow$). 
The evaluation is conducted on multiple datasets including \textbf{Flare-R} from Flare7K++ for deflare, 
\textbf{UHDM} for moiré pattern removalremoval, and the \textbf{SIR$^2$+} reflection removal benchmark with three subsets: 
PostcardDataset, SolidObjectDataset, and WildScene. 
Flare-R contains real captured flare images, while UHDM provides ultra-high-definition paired moiré images. 
The SIR$^2$+ subsets represent different reflection scenarios with diverse scene contents. 
The best results are highlighted in \textbf{bold}, and the second-best results are \underline{underlined}.
}
\label{tab:extra_comparison}
\resizebox{\textwidth}{!}{
\begin{tabular}{l | cc | cc | cc | cc | cc | cc}
\toprule
\multirow{2}{*}{\textbf{Method}} 
& \multicolumn{2}{c|}{\textbf{Flare}} 
& \multicolumn{2}{c|}{\textbf{Moiré}} 
& \multicolumn{2}{c|}{\textbf{Refl (Postcard)}} 
& \multicolumn{2}{c|}{\textbf{Refl (Solid)}} 
& \multicolumn{2}{c|}{\textbf{Refl (Wildscene)}} 
& \multicolumn{2}{c}{\textbf{Average}} \\
\cmidrule(lr){2-3} \cmidrule(lr){4-5} \cmidrule(lr){6-7} \cmidrule(lr){8-9} \cmidrule(lr){10-11} \cmidrule(lr){12-13}
 & PSNR$\uparrow$ & SSIM$\uparrow$ & PSNR$\uparrow$ & SSIM$\uparrow$ & PSNR$\uparrow$ & SSIM$\uparrow$ & PSNR$\uparrow$ & SSIM$\uparrow$ & PSNR$\uparrow$ & SSIM$\uparrow$ & PSNR$\uparrow$ & SSIM$\uparrow$ \\
\midrule
Nano Banana Pro & \textbf{25.28} & \textbf{0.889} & \textbf{18.86} & \underline{0.726} & 20.11 & 0.821 & \textbf{24.53} & \textbf{0.880} & \underline{22.27} & 0.835 & \textbf{22.21} & \textbf{0.830} \\
GPT\textendash Image\textendash 1.5 & 16.86 & 0.526 & 11.84 & 0.549 & 13.51 & 0.465 & 13.97 & 0.451 & 14.59 & 0.527 & 14.15 & 0.504 \\
Seedream 4.5 & \underline{23.88} & 0.831 & 15.30 & 0.643 & \underline{21.38} & 0.807 & \underline{22.16} & 0.809 & 21.45 & 0.783 & 20.83 & 0.775 \\
LongCat\textendash Image\textendash Edit & 22.57 & 0.809 & 15.79 & 0.670 & 20.28 & 0.706 & 17.12 & 0.632 & 21.28 & 0.805 & 19.41 & 0.724 \\
Qwen\textendash Image\textendash Edit\textendash 2511 & 22.45 & \underline{0.869} & 16.07 & 0.673 & 21.07 & 0.839 & 20.17 & \underline{0.864} & \textbf{23.77} & \textbf{0.896} & 20.71 & \underline{0.828} \\
FLUX.1-Kontext-dev & 23.27 & 0.859 & 10.36 & 0.516 & 13.43 & 0.428 & 11.60 & 0.374 & 14.62 & 0.520 & 14.66 & 0.539 \\
Step1X\textendash Edit & 18.74 & 0.772 & 14.31 & 0.583 & 19.29 & \underline{0.841} & 18.56 & 0.783 & 21.12 & \underline{0.864} & 18.40 & 0.769 \\
\midrule
\textbf{Ours} & 22.05 & 0.745 & \underline{17.62} & \textbf{0.743} & \textbf{22.67} & \textbf{0.904} & 20.35 & 0.797 & 21.72 & 0.826 & \underline{20.88} & 0.803 \\
\bottomrule
\end{tabular}}
\end{table*}

\section{Ablation Study Details}
Besides comparing the proposed two-stage training strategy composed of the synthetic degradation transfer training stage and the real-world degradation SFT stage, we further analyze an alternative setting where the model is trained using only real-world degradation data. For a fair comparison, we train the model for the same number of iterations as in the synthetic transfer training stage. At the peak point, the model trained only on real-world data tends to overfit the degradation patterns, which harms the structural consistency of the restored images. This often results in artifacts such as object deformation, body shifting, and unrealistic enhancement. These observations further confirm the importance of the proposed two-stage training strategy.

Specifically, at 2.5K training steps, the model trained with only synthetic degradation data still shows limited ability to handle complex real-world degradations, while the model trained solely on real-world degradation data can partially restore degradations but often fails to preserve content consistency. The model produces overly enhanced results, such as removing natural light sources, as shown in Figure~\ref{fig:ablation}. In contrast, our two-stage training strategy effectively balances restoration capability and structural consistency, leading to more stable and generalizable performance.

\section{User Study Details} \label{sec:ablation}
\begin{figure*}[htbp]
  \centering
  \setlength{\abovecaptionskip}{2pt}
  \includegraphics[width=0.98\linewidth]{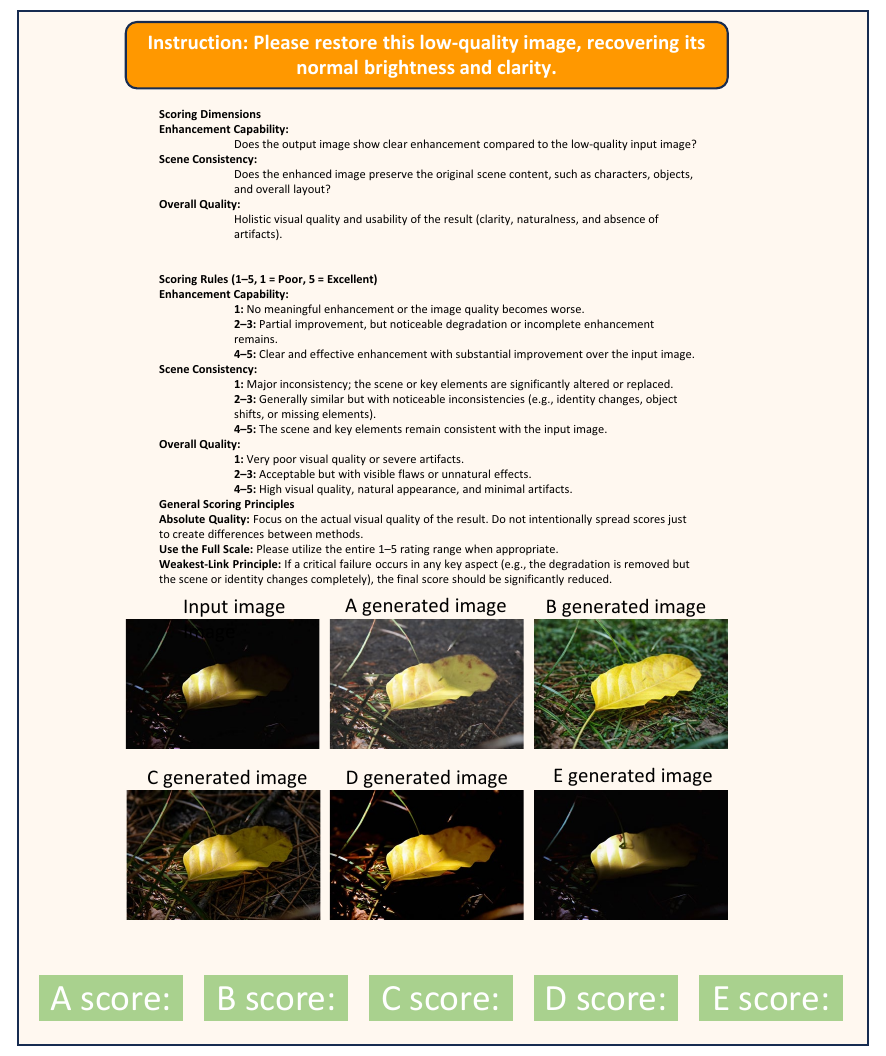}
  \vspace{-0.5ex}
  \caption{\small User study interface used to evaluate the restoration results. Participants are presented with one degraded input image and five restored results generated by different models and are asked to rate them based on restoration quality and consistency.}
  \vspace{-3ex}
  \label{fig:user_study}
\end{figure*}

Our user study evaluates the results of the five best-performing models on \textbf{RealIR-Bench}. All the 32 participants receive a brief tutorial beforehand to ensure they understand the task and the evaluation criteria. The interface used in the user study is illustrated in Figure~\ref{fig:user_study}. To further analyze the reliability of the proposed metric, we compute the Kendall’s $\tau_b$, Spearman Rank Correlation Coefficient (SRCC), and Pearson Linear Correlation Coefficient (PLCC) between the metric score (FS) and human judgments. These correlation measures are widely used to evaluate the consistency between automatic metrics and human evaluation. The results demonstrate that the proposed metric achieves moderate statistical alignment with human judgments ($p < 0.01$) across all evaluation settings, as shown in Table~\ref{tab:consistency_metrics}.
% usee
\begin{table}[h]
\centering
\caption{Consistency evaluation between RS metric and subjective human perception.}
\label{tab:consistency_metrics}
\begin{tabular}{lcc}
\toprule
\textbf{Metric} & \textbf{Correlation Coefficient} & \textbf{$p$-value} \\
\midrule
Kendall's $\tau_b$ & 0.2493 & $2.96 \times 10^{-124}$ \\
SRCC               & 0.3010 & $4.62 \times 10^{-128}$ \\
PLCC               & 0.2919 & $3.21 \times 10^{-120}$ \\
\bottomrule
\end{tabular}
\end{table}

\clearpage

% fig: training data

\begin{figure*}[htbp]
    \centering
    \setlength{\abovecaptionskip}{2pt}
    \includegraphics[width=0.98\linewidth]{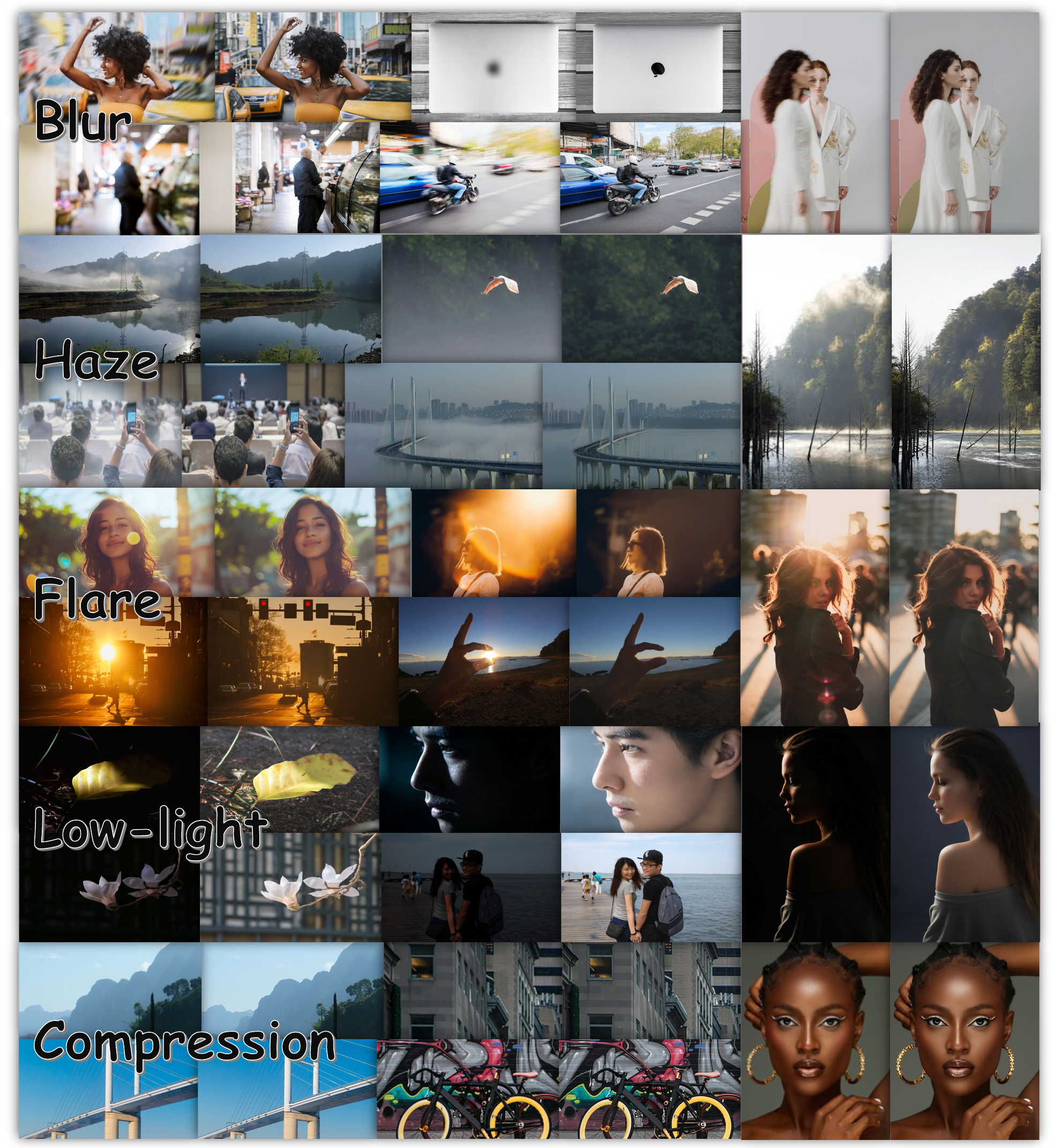}
    \vspace{-0.5ex}
    \caption{\small Additional qualitative results of RealRestorer under real-world degradations. Please zoom in for better visualization of details.}
    \vspace{-3ex}
    \label{fig:finalshow}
\end{figure*}

\begin{figure*}[htbp]
    \centering
    \setlength{\abovecaptionskip}{2pt}
    \includegraphics[width=0.98\linewidth]{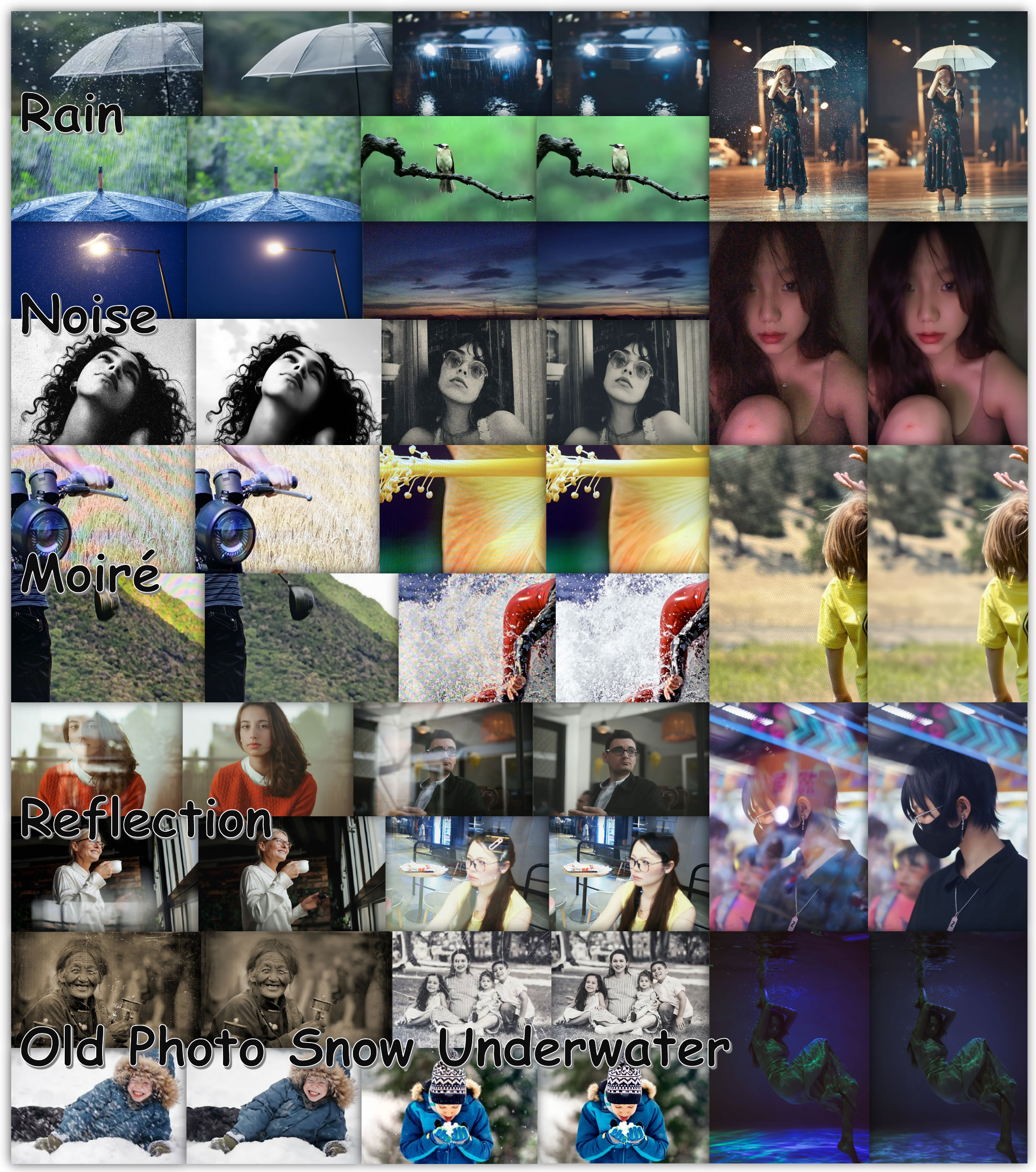}
    \vspace{-0.5ex}
    \caption{\small Additional qualitative results of RealRestorer under real-world degradations. Please zoom in for better visualization of details.}
    \vspace{-3ex}
    \label{fig:finalshow2}
\end{figure*}

% fig:benchmark

\clearpage

% table_no_caption: vlm instruction

% \begin{wrapfigure}{r}{1\textwidth}
% \centering
% \vspace{35mm}
% \begin{tcolorbox}[
% title=Instruction for Degradation Evaluation,
% enhanced,
% colback={rgb,255:red,252;green,247;blue,238},
% colframe={rgb,255:red,196;green,132;blue,58},
% fonttitle=\bfseries,
% width=\linewidth
% ]

% \begin{itemize}
% \item \textbf{Degradation Evaluation:} You will evaluate whether an image exhibits the degradation type \{task\} and how severe that degradation is. The evaluation should focus only on the specified degradation type and ignore unrelated image-quality issues and semantic content.

% \item \textbf{Evaluation Procedure:}
% \begin{enumerate}
% \item Divide the image into regions (e.g., a $3\times3$ grid).
% \item Inspect each region for the presence and severity of \{task\}.
% \item Estimate the proportion of affected areas and the severity.
% \item Aggregate regional observations into a single score.
% \item Weight the score by the affected area if degradation is uneven.
% \item For borderline cases, assign the nearest score level.
% \item Do not output any reasoning.
% \end{enumerate}

% \item \textbf{Scoring Scale (1--5):}
% \begin{itemize}
% \item 5 = No \{task\}
% \item 4 = Mild ($\leq20\%$ area)
% \item 3 = Moderate (20--50\%)
% \item 2 = Severe (50--80\%)
% \item 1 = Extreme ($>80\%$)
% \end{itemize}

% \item \textbf{Output Format:} Return only ``Degradation Score: <1--5>''.
% \end{itemize}

% \end{tcolorbox}

% \caption{System instruction used for degradation evaluation.}
% \label{fig:degradation_instruction}

% \end{wrapfigure}
\begin{figure*}[t]
\centering
\begin{tcolorbox}[
    title=Instruction for Degradation Evaluation,
    enhanced,
    colback={rgb,255:red,252;green,247;blue,238},
    colframe={rgb,255:red,196;green,132;blue,58},
    fonttitle=\bfseries\Large,
    fontupper=\large,
    width=0.9\textwidth,
    boxsep=6pt,
    left=10pt, right=10pt,
    top=10pt, bottom=10pt,
    height=0.78\textheight,
]

\setlength{\itemsep}{10pt}
\setlength{\parsep}{6pt}

\begin{itemize}
\item \textbf{Degradation Evaluation:} You will evaluate whether an image exhibits the degradation type \{task\} and determine its severity level. The assessment should focus exclusively on the specified degradation type, while ignoring unrelated image-quality issues, semantic content, and aesthetic preference. Your goal is to provide a consistent and objective judgment of how strongly the target degradation affects the image.

\item \textbf{Evaluation Procedure:}
\begin{enumerate}
\item Divide the image into several local regions (e.g., a $3\times3$ grid) to ensure a systematic inspection.
\item Examine each region carefully and identify whether the degradation type \{task\} is present.
\item Estimate both the severity of the degradation and the proportion of the image area that is affected.
\item Summarize the regional observations into a single overall score that best reflects the image-level degradation.
\item If the degradation is spatially uneven, place greater emphasis on the regions that are more strongly affected.
\item For borderline cases, choose the nearest score level based on the dominant visual impression.
\item Do not output any intermediate reasoning, explanation, or analysis.
\end{enumerate}

\item \textbf{Scoring Scale (1--5):} Use the following scale to measure the severity of the target degradation. Higher scores indicate cleaner images with less visible degradation, while lower scores indicate stronger and more widespread corruption.
\begin{itemize}
\item 5 = No \{task\}; the image is essentially clean with no noticeable degradation.
\item 4 = Mild degradation; only a small portion of the image is affected ($\leq 20\%$ area), and the overall visual quality remains largely intact.
\item 3 = Moderate degradation; the degradation is clearly visible and affects a noticeable part of the image (20--50\% area).
\item 2 = Severe degradation; the corruption is strong and influences a large portion of the image (50--80\% area).
\item 1 = Extreme degradation; the degradation dominates most of the image ($> 80\%$ area) and seriously harms visual quality.
\end{itemize}

\item \textbf{Output Format:} Return only in the format ``Degradation Score: <1--5>''.
\end{itemize}

\end{tcolorbox}
\caption{\small System instruction used for degradation evaluation.}
\label{fig:degradation_instruction}
\end{figure*}

\begin{figure*}[htbp]
    \centering
    \setlength{\abovecaptionskip}{2pt}
    \includegraphics[width=0.98\linewidth]{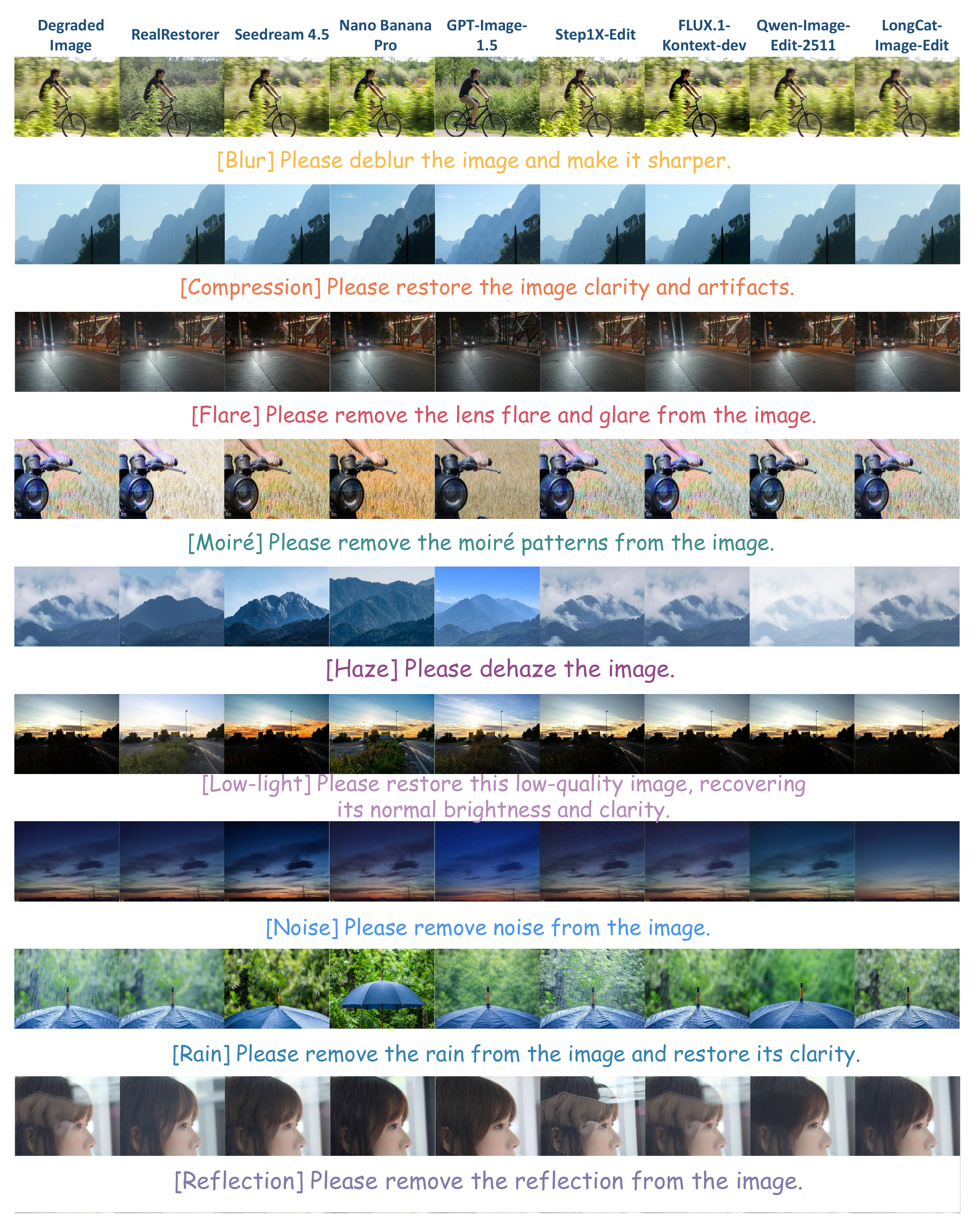}
    \vspace{-0.5ex}
    \caption{\small More qualitative comparison results on RealIR-Bench. Zoom in to see more details.}
    \vspace{-3ex}
    \label{fig:comparision1}
\end{figure*}

\begin{figure*}[htbp]
    \centering
    \setlength{\abovecaptionskip}{2pt}
    \includegraphics[width=0.98\linewidth]{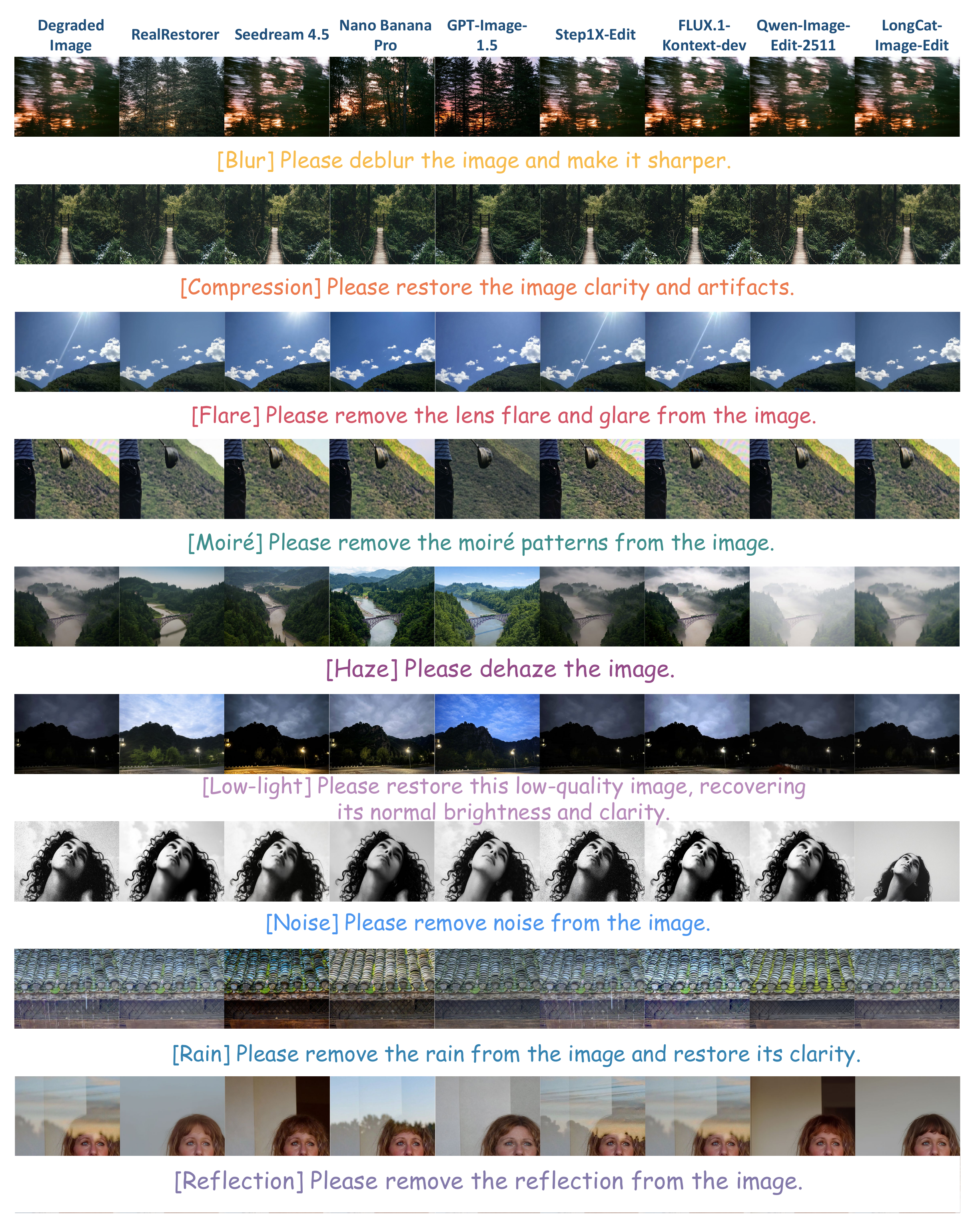}
    \vspace{-0.5ex}
    \caption{\small More qualitative comparison results on RealIR-Bench. Zoom in to see more details.}
    \vspace{-3ex}
    \label{fig:comparision2}
\end{figure*}

\begin{figure*}[htbp]
    \centering
    \setlength{\abovecaptionskip}{2pt}
    \includegraphics[width=0.98\linewidth]{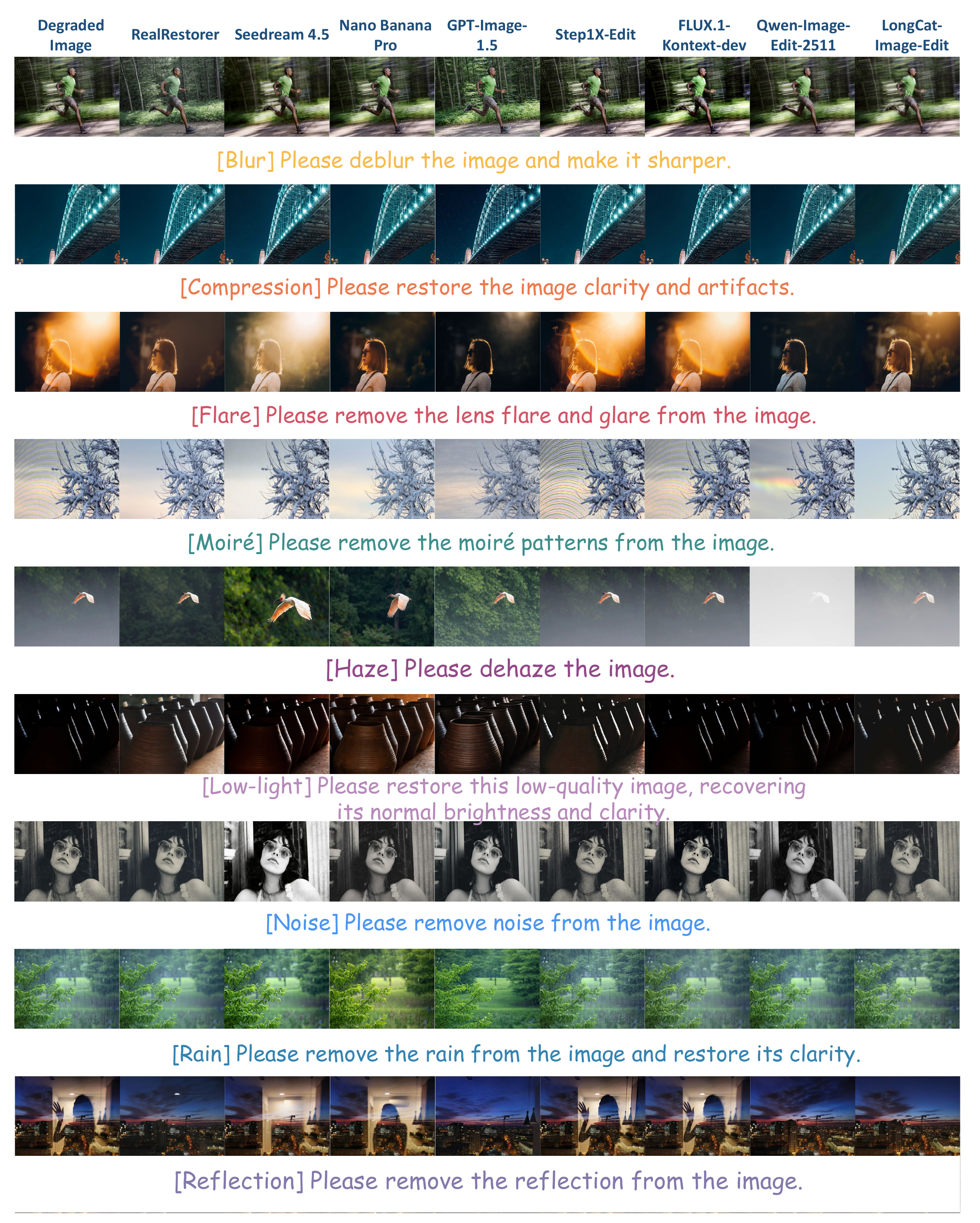}
    \vspace{-0.5ex}
    \caption{\small More qualitative comparison results on RealIR-Bench. Zoom in to see more details.}
    \vspace{-3ex}
    \label{fig:comparision3}
\end{figure*}

\begin{figure*}[htbp]
    \centering
    \setlength{\abovecaptionskip}{2pt}
    \includegraphics[width=0.98\linewidth]{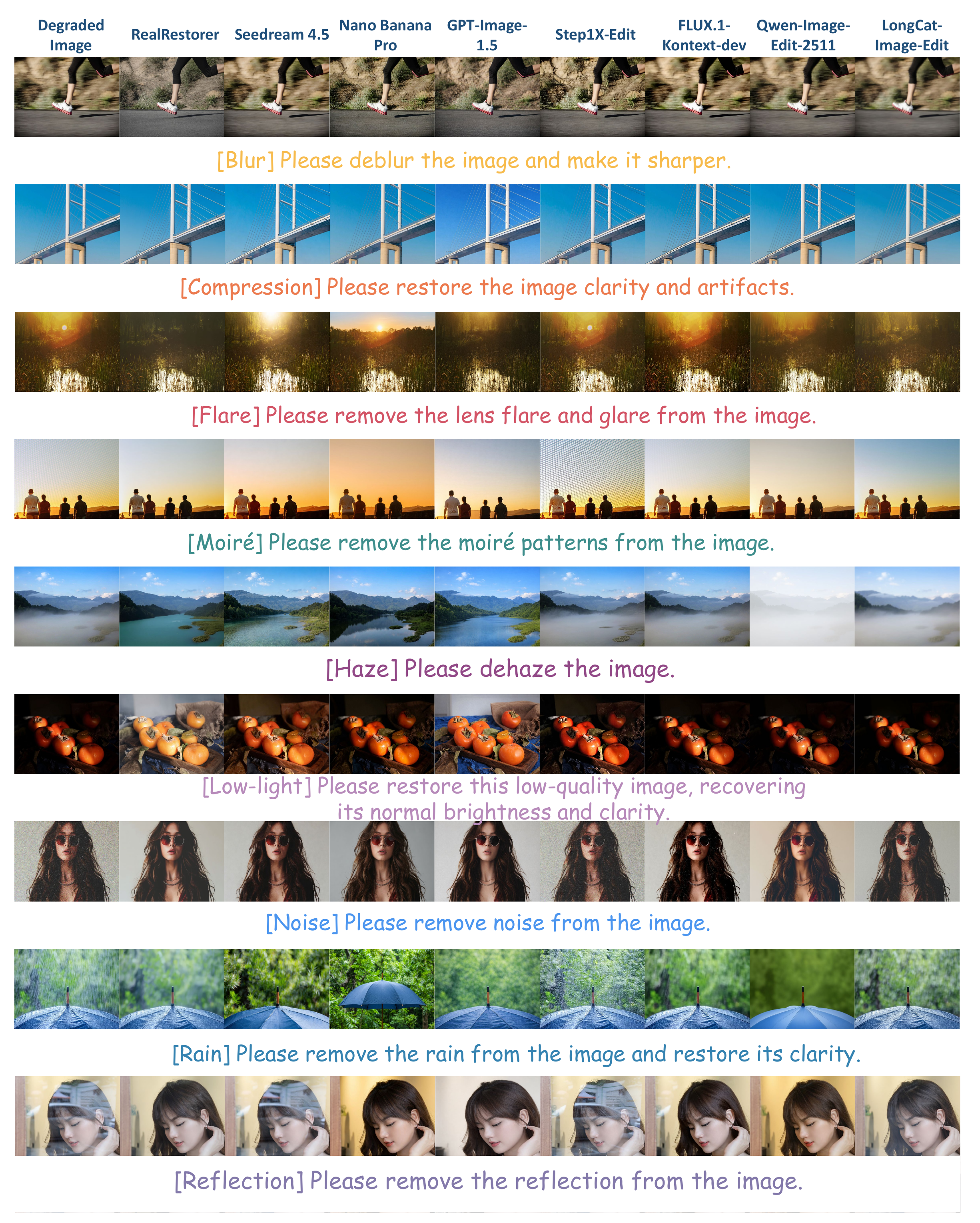}
    \vspace{-0.5ex}
    \caption{\small More qualitative comparison results on RealIR-Bench. Zoom in to see more details.}
    \vspace{-3ex}
    \label{fig:comparision4}
\end{figure*}

% ==========================================
% Ablation Study
% ==========================================

\begin{figure*}[htbp]
  \centering
  \setlength{\abovecaptionskip}{2pt}
  \includegraphics[width=0.98\linewidth]{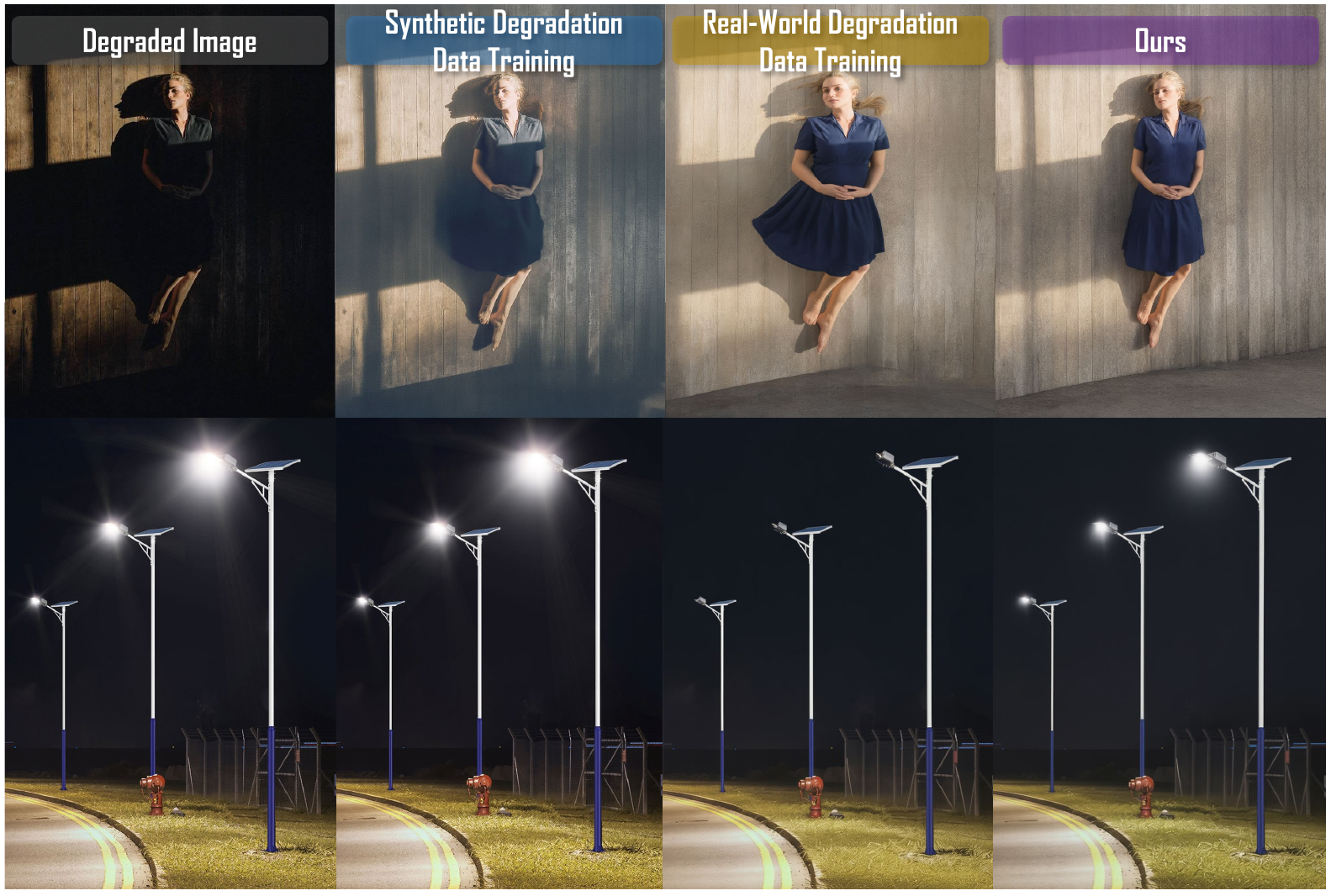}
  \vspace{-0.5ex}
  \caption{\small Qualitative comparison of different training strategies. Models trained only with synthetic degradation data show limited ability to restore complex real-world degradations. In contrast, models trained solely on real-world degradation data tend to overfit, which may harm structural consistency. Our two-stage training strategy effectively balances restoration capability and content consistency.}
  \vspace{-3ex}
  \label{fig:ablation}
\end{figure*}

% ==========================================
% User Study
% ==========================================

\end{document}